\documentclass[runningheads]{llncs}

\usepackage{eccv}

\usepackage{eccvabbrv}
\usepackage{amsmath}
\usepackage{algorithmic}
\usepackage{algorithm}

\usepackage{graphicx}
\usepackage{booktabs}

\usepackage{colortbl}
\usepackage{xcolor}
\usepackage{multicol}
\usepackage{multirow}

\usepackage{subcaption}

\usepackage[accsupp]{axessibility}  

\usepackage[utf8]{inputenc}
\usepackage{algorithm}
\usepackage{algorithmic}
\usepackage{tikz}
\usepackage{booktabs}
\usepackage{multirow}
\usepackage{xcolor} 
\usepackage[table]{xcolor}

\definecolor{sameview}{RGB}{66,132,156}
\definecolor{targetview}{RGB}{247,150,150}

\def\1{\bm{1}}

\DeclareMathAlphabet{\mathsfit}{\encodingdefault}{\sfdefault}{m}{sl}
\SetMathAlphabet{\mathsfit}{bold}{\encodingdefault}{\sfdefault}{bx}{n}
\usepackage[utf8]{inputenc}
\usepackage{algorithm}
\usepackage{algorithmic}
\usepackage{tikz}
\usepackage{booktabs}
\usepackage{multirow}

\usepackage{hyperref}
\usepackage{graphicx}
\usepackage{booktabs}
\usepackage{colortbl}
\usepackage{xcolor}
\usepackage{multirow}
\usepackage{bbm}
\usepackage{dsfont}

\usepackage{orcidlink}

\begin{document}

\title{VisTa3D: A Dataset and Benchmark for \\
Thin Object Reconstruction from \\ 
\underline{Vis}ion, \underline{Ta}ctile, and \underline{3D} Point Clouds} 

\titlerunning{Abbreviated paper title}

\author{
    Shania Guo\inst{1}\orcidlink{0009-0006-9159-3607}\index{Guo, Shania} \and
    Yeongsik Seo\inst{2}\orcidlink{0000-0003-2423-2167}\index{Seo, Yeongsik} \and
    Andrew Fu\inst{1}\orcidlink{0009-0002-9920-9121} \and
    Mei Hao\inst{2}\orcidlink{0000-0002-4287-6794}\index{Hao, Mei} \and
    Iris Xia\inst{1} \and \\
    Jiwon Jenny Lee\inst{1}\orcidlink{0009-0006-1435-4418}\index{Lee, Jiwon}  \and 
    Xinyi Mary Xie\inst{1}\orcidlink{0009-0000-6538-0794}\index{Xie, Xinyi Mary}  \and
    Hyoungseob Park\inst{1}\orcidlink{0000-0003-0787-2082} \and \\
    Aaron Dollar\inst{2}\orcidlink{0000-0002-2409-4668}\index{Dollar, Aaron} \and
    Alex Wong\inst{1} \orcidlink{0000-0002-3157-6016}\index{Wong, Alex}
}

\authorrunning{S.~Guo et al.}
\titlerunning{VisTa3D}

\institute{
    Yale Vision Laboratory, Yale University, New Haven, CT 06520, USA \\
    \email{\{shania.guo,andrew.fu,jiwon.jenny.lee,x.xie,hyoungseob.park,alex.wong\}@yale.edu}
\and
    GRAB Lab, Yale University, New Haven, CT 06520, USA \\
\email{\{yeongsik.seo,mei.hao,aaron.dollar\}@yale.edu}
}

\maketitle

\begin{abstract}

State-of-the-art 3D reconstruction models, whether from visual, range, or both, tend to underperform on thin objects. This is partially due to the small amount of space such objects occupy in RGB images and in 3D point clouds. To test the extent of their errors, we collected the first thin object dataset comprising synchronized RGB images, depth maps, and tactile response maps, where each frame is associated with inertial measurements, camera pose and calibration, and ground-truth depth and segmentation maps obtained from laser scanning of thin objects. We hypothesize that tactile data can aid in the reconstruction of thin objects as their response maps provide local shape and deformation information. Our dataset, termed \textbf{VisTa3D}, comprises of 387 scenes covering 70 thin objects over 17 environments. We benchmarked current 3D reconstruction models on VisTa3D and found that, indeed, they exhibit low fidelity on thin objects. To test if tactile data can help, we introduce the first visual-range-tactile 3D reconstruction model as a baseline. Code and data: \url{https://huggingface.co/datasets/shaniaguo/VisTa3D}.

\end{abstract}

\section{Introduction}
\label{sec:intro}

\begin{figure*}[t]
    \centering
   \includegraphics[width=\linewidth]{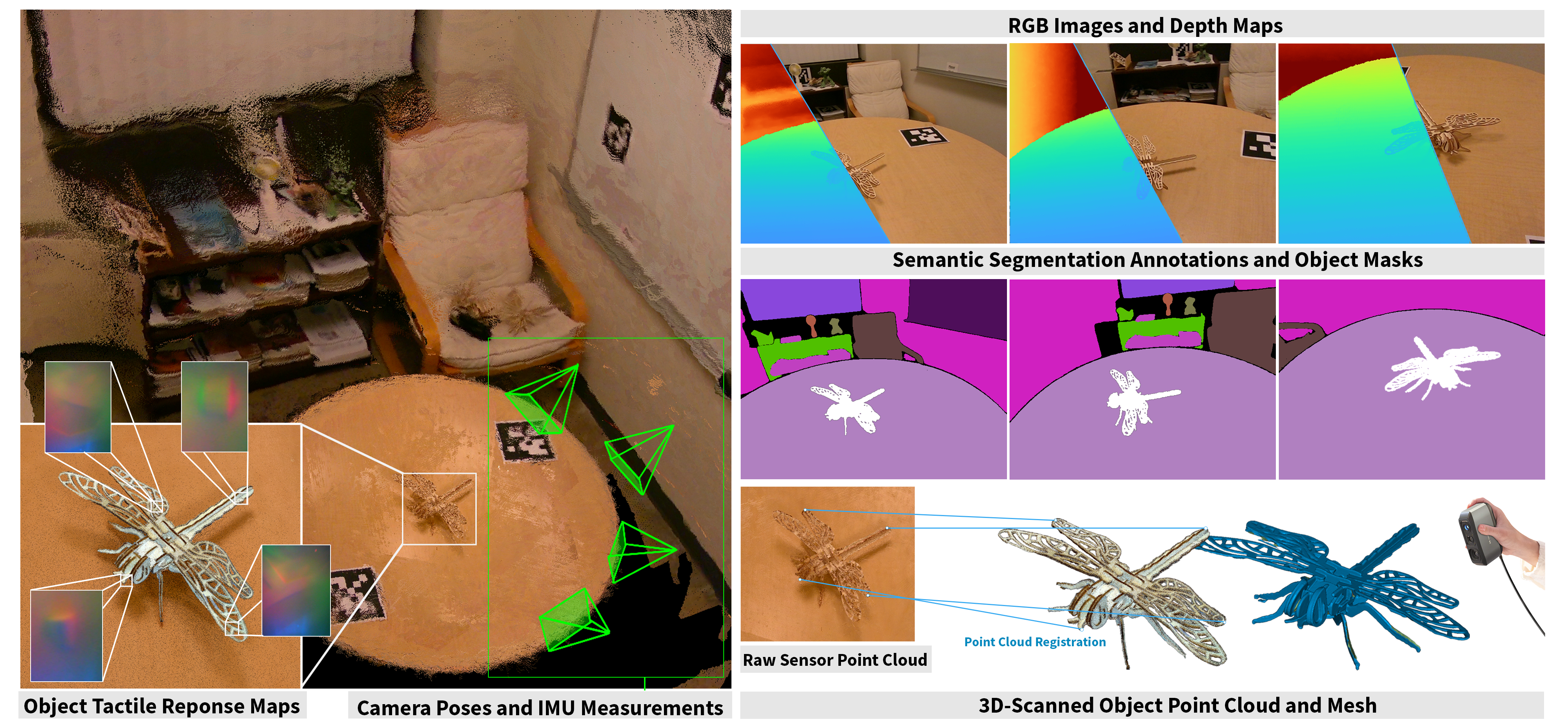}
   \vspace{-6mm}
    \caption{\textit{VisTa3D} comprises synchronized RGB images, depth maps, tactile response maps, and inertial measurements with camera poses, and ground truth depth maps and segmentation masks derived from 3D scanned object point clouds and meshes.}
    \label{fig:teaser}
    \vspace{-2mm}
\end{figure*}

Three-dimensional (3D) reconstruction of thin objects from visual and range data is challenging. The projection of thin objects onto the RGB image only occupies a small number of pixels; likewise, their returns from range sensors also tend to be sparse. This is partially due to the limited space such objects occupy in 3D as well as the finite resolution and sampling of commercial sensors, especially at longer distances. As such, even small sensor motions may cause parts of the object to be occluded (or disoccluded), which leads to a loss of correspondence across viewpoints and thus degrades reconstruction quality. Although high-resolution sensors may better capture thin objects, they are impractical in cost and require larger computational resources during downstream processing. 

Recent 3D reconstruction models, trained on large-scale datasets, have generalized well across diverse 3D scenes. 
Despite the wide range of existing benchmark datasets, none focus on thin structures. To test the extent in which thin objects pose a challenge to current 3D reconstruction models, e.g., monocular depth estimation, multi-view stereo, depth completion and neural rendering methods, we collected a dataset of diverse real-world objects in both indoor and outdoor environments. We found that these models exhibit pronounced errors on thin structures. We hypothesize this is due to the limited representation of thin objects in existing training sets, as well as the limited space they occupy.

While commercial visual and range or depth sensors often fail to capture thin structures, tactile sensing offers local measurements of surface shape and deformation that are largely independent of object thickness. Tactile sensors are also complementary to visual and range sensors. At close distances, occlusion becomes more prominent and can obstruct visual observation of thin objects, while the object may also fall below the minimum sensing range of depth sensors. In contrast, tactile sensing remains effective under these conditions. 

To test the potential of tactile sensing in conjunction with visual and range inputs, we introduce the first visual–range–tactile dataset for 3D reconstruction of thin objects. Our dataset includes synchronized RGB images, depth maps, and tactile response maps with inertial measurements, camera poses and calibration, and ground-truth depth maps and segmentation masks. To capture the intricate details of thin objects, we use a high-fidelity laser scanner to reconstruct each object as a 3D digital asset. We use a commercial sensor for raw data capture, which is aggregated and refined. Then, we register and insert the digital assets into the reconstruction to yield ground truth. As a byproduct, this also produces high-fidelity segmentation masks corresponding to each thin object. We organize our dataset as a benchmark aimed to test performance on thin objects. We further collect a synthetic dataset of the same data modalities to seed the (pre)training of methods for this benchmark. Fig.~\ref{fig:teaser} shows an overview of the proposed dataset and benchmark of \textbf{Vis}ion, \textbf{Ta}ctile, and \textbf{3D} point clouds for thin object reconstruction, or \textbf{VisTa3D} for short. Our dataset comprises 387 scenes covering 70 thin objects over 17 environments, both real and synthetic. 

As a step towards enabling high-fidelity reconstruction through the fusion of visual, tactile, and range modalities, we further propose the first (baseline) model that incorporates all three modalities. Inspired by design of depth completion networks, we found that a naive integration of tactile into visual-range fusion improves on thin object reconstruction when trained only with the synthetic portion of VisTa3D, and even more when trained on synthetic and real data. 

\textbf{Our contributions}: (1) We introduce the first multimodal dataset for thin object reconstruction comprising synchronized RGB images, depth maps, and tactile response maps with inertial measurements, camera calibration and poses, and associated ground-truth depth maps and segmentation masks. (2) We propose an extensive data curation pipeline that involves high-fidelity laser scanning of thin objects and postprocessing that derives the above quantities. (3) We benchmark 11 existing methods on our dataset under 6 evaluation modes and found that they consistently struggle on thin objects.
(4) To address this, we propose the first vision-tactile-range 3D reconstruction model as a baseline.

\section{Related Work}

Recent advances in 3D reconstruction, e.g., neural rendering, feed-forward models, have achieved high-fidelity outputs; yet, thin structures remain challenging.

\textbf{Monocular Depth Estimation} (MDE) \cite{eigen2015predicting,fei2019geo,laina2016deeper,garg2016unsupervised,gangopadhyay2025extending,gangopadhyay2026from,godard2017unsupervised,godard2019digging,lao2024depth,lao2024sub,upadhyay2023enhancing,wong2019bilateral,wu2024augundo} aims to infer a depth map from a single image. Recent works have leveraged large-scale datasets to train foundation models \cite{ranftl2020towards,ranftl2021vision,piccinelli2024unidepth,piccinelli2025unidepthv2,yang2024depthanythingv1,yang2024depthanythingv2}, capable of generalizing across diverse 3D scenes, and have set the state of the art. \cite{ranftl2020towards} demonstrated generalizable MDE through mixed-dataset training and utilized a gradient-matching loss to preserve fine details. ~\cite{ranftl2021vision} introduced a Transformer-based dense decoder. DepthAnythingV1 \cite{yang2024depthanythingv1} proposed to train on pseudolabels generated by a pretrained seed model. DepthAnythingV2 \cite{yang2024depthanythingv2} extended \cite{yang2024depthanythingv1} to using synthetic data to train the seed model to capture finer details. UniDepth V1 \cite{piccinelli2024unidepth} introduced a pseudo-spherical representation to generalize across cameras and utilized an edge-guided loss to maintain sharp transitions; UniDepth V2 \cite{piccinelli2025unidepthv2} simplified the architecture of \cite{piccinelli2024unidepth}. 
Yet, little work has benchmarked MDEs on thin structures.

\textbf{Multi-View Stereo} (MVS) \cite{duan2026fisheye3r,gu2020cascade,romanoni2019tapa,wang2021patchmatchnet,wang2024dust3r,wang2024multiview,yao2018mvsnet} reconstructs the 3D scene from multiple images. Traditional MVS methods employing photometric consistency struggle fundamentally with thin objects due to correspondence ambiguity~\cite{xu2020marmvs}. \cite{romanoni2019tapa,wang2024multiview} recognize thin structures as persistent challenges where matching ambiguities arise.
Learning-based methods have evolved from cost volume approaches~\cite{yao2018mvsnet} to cascade formulations enabling high-resolution reconstruction~\cite{gu2020cascade}. \cite{wang2021patchmatchnet} proposed adaptive spatial cost aggregation for boundary stability. \cite{zhang2023geomvsnet} fused feature across resolutions by multiple branches. \cite{wang2024dust3r} introduced the first feedforward 3D reconstruction model. Foundation models \cite{keetha2026mapanything,wang2025vggt} follow \cite{wang2024dust3r} and extend to large-scale datasets to generalize across diverse 3D scenes.  VGGT~\cite{wang2025vggt} leverages global attention in transformers for feed-forward 3D inference; MapAnything \cite{keetha2026mapanything} introduces additional modalities by tokenizing inputs. Neither have been specifically evaluated on thin objects. 

\textbf{Monocular Depth Completion} (MDC) \cite{ezhov2024all,kam2022costdcnet,li2020multi,lin2022dynamic,park2020non,park2026orcas,liu2022monitored,rim2025protodepth,wong2021adaptive,wong2021learning,wong2021unsupervised,zhang2023completionformer,zuo2025omni} infers a dense depth map from an image and sparse depth map. 
Depth completion is particularly challenging for thin objects as few points will be returned by a range sensor. CPSN \cite{cheng2019learning,cheng2020cspn++} proposed spatial propagation. NLSPN \cite{park2020non} extended \cite{cheng2020cspn++} to non-local propagation while DySPN \cite{lin2022dynamic} to dynamic propagation. VOICED \cite{wong2020unsupervised}  and BPNet \cite{tang2024bilateral} prefilled sparse depth map with scaffolding and bilateral propagation. \cite{chung2025eta,park2024test} proposed test-time adaptation while \cite{rim2026radar,singh2023depth} performed completion on extremely sparse radar points. \cite{kam2022costdcnet} fused image and sparse depth via a cost volume. \cite{zhang2023completionformer} leveraged local and global context with a transformer. OMNI-DC~\cite{zuo2025omni} achieved zero-shot generalization across sparse depth patterns through multi-resolution depth integration with predicted depth maps from foundation monocular depth estimators as guidance.

\textbf{Novel View Synthesis} (NVS)
\cite{mildenhall2021nerf,barron2021mip,barron2022mip, barron2023zip,duan2026evidential,kerbl20233d, huang20242d,wang2026ode} reconstructs 3D scenes from multiple images. 
NeRF~\cite{mildenhall2021nerf} introduced coordinate-based neural radiance fields optimized through differentiable volume rendering. \cite{barron2021mip} and \cite{barron2022mip} addressed aliasing with multiscale representations, while ~\cite{muller2022instant} accelerated optimization using multiresolution hash encodings. \cite{barron2023zip}  combined grid-based representations with multi-sampling-based prefiltering to reduce z-aliasing, where thin structures may disappear. However, volumetric neural representations can still blur or miss thin structures due to lack of correspondence, as they tend to be occluded (or disoccluded) from view with small camera motion. 
3D Gaussian Splatting~\cite{kerbl20233d} introduced explicit anisotropic Gaussian primitives for real-time rendering, but volumetric Gaussians remain prone to self-occlusion artifacts around thin structures. 2D Gaussian Splatting~\cite{huang20242d} represented scenes with oriented planar disks to improve view-consistent surface modeling, yet extremely thin structures remain challenging when optimization provides weak or inconsistent constraints.

\textbf{Tactile Sensing and Reconstruction} \cite{abad2020visuotactile,bauza2019tactile, comi2024touchsdf,dou2024tactile,huang2025gelslam,lu2023tac2structure,suresh2022shapemap,suresh2024neuralfeels,swann2024touch,tu2026unitac,wang20183d,yang2024binding,zhao2023fingerslam,zhu2025forces} use contact-based local surface measurements, e.g., vision-based tactile sensors~\cite{yuan2017gelsight}, for reconstruction. Prior works have used tactile sensing as a standalone modality~\cite{lu2023tac2structure, comi2024touchsdf, huang2025gelslam} or combined it with complementary signals such as vision~\cite{zhao2023fingerslam, wang20183d}, depth~\cite{suresh2022shapemap}, and robot kinematics~\cite{suresh2024neuralfeels, bauza2019tactile}. 
TouchSDF~\cite{comi2024touchsdf} proposed an implicit signed distance representation. ~\cite{dou2024tactile} unified visual and tactile observations within a shared radiance field. Gaussian-splatting approaches incorporate tactile cues into explicit scene representations: ~\cite{swann2024touch} supervised 3D Gaussian Splatting with fused tactile depth and uncertainty maps, while \cite{comi2025snap} injected local tactile responses into Gaussian optimization. These methods show that tactile data can provide reliable local information for neural 3D representations, but their effectiveness for thin-object reconstruction remains underexplored.

\textbf{3D Reconstruction of Thin Structures}
remains challenging due to insufficient sampling, lack of correspondences across views, and self-occlusion (e.g., tubular objects and tree branches). 
\cite{hilton1996reconstruction,pulli1997robust,hilton1998implicit,bornik2005reconstruction,ummenhofer2013point,tabb2013shape,martin2014topology,yucer2016depth,li2018reconstructing,liu2021curvefusion,wang2020vid2curve,chou2022gensdf} study reconstruction methods for thin surfaces, curves, and tubular structures. ~\cite{hilton1996reconstruction} reconstructs thin surface regions, ~\cite{pulli1997robust} provide robust mesh representations, and ~\cite{hilton1998implicit} better preserve thin surface sections. Other approaches exploit structural priors for thin features. ~\cite{bornik2005reconstruction} reconstruct tubular objects through contour connections, ~\cite{martin2014topology} targets thin tubular structures, ~\cite{li2018reconstructing} recover thin manifold surfaces, ~\cite{liu2021curvefusion} reconstruct filament-like structure from dense depth, and ~\cite{wang2020vid2curve} estimates continuous 3D curves and radii. Methods to improve thin-object reconstruction include multi-view depth fusion~\cite{ummenhofer2013point}, silhouette probability maps~\cite{tabb2013shape}, densely sampled light fields~\cite{yucer2016depth}, and signed nearest-neighbor losses for SDF learning~\cite{chou2022gensdf}.

\textbf{Thin Object Datasets.}
Existing datasets on thin structures predominantly target 2D image segmentation.
ThinObject-5K~\cite{liew2021deep} provides large-scale annotations for thin object segmentation (5,743 objects), including bicycle spokes, wires, and cables. For powerline segmentation, Mendeley Powerline ~\cite{yetgin2017powerline} offers paired visible-light and infrared (IR) images, while PLDU and PLDM~\cite{zhang2019detecting} provide UAV-captured RGB images with dense segmentation labels. Beyond image segmentation, Zagar \textit{et al.}~\cite{vzagar20253d} introduce point cloud datasets for thin structures, PointWire (real-world wiring harness scans) and PointVessel (synthetic tubular vessel geometry); however, they are limited to only point clouds.

In contrast, we introduce a \emph{multimodal 3D thin object reconstruction} benchmark with visual, range, inertial and tactile data that are associated with camera poses and calibration, both real and synthetic. We provide high-fidelity ground-truth depth maps and segmentation masks derived from 3D-scanned objects. We evaluate state-of-the-art 3D reconstruction methods on our dataset and show that existing methods remain limited on thin structures.

\section{Data Collection Methodology}
Our data collection platform includes an Intel RealSense D455 for RGB images, depth maps, and inertial measurements, DIGIT tactile sensor for vision-based tactile response maps, and a 3DMakerPro Moose, a structured-light 3D scanner, for high-precision point clouds. Note: RGB images and depth maps share the same resolution; depth maps are registered to RGB image frame.  

\textbf{Vision, Inertial, Range Data Collection.} We set up each scene by placing a single object on a support surface, e.g., table, bench, in a selected environment. Each sequence comprise of a single trajectory around the scene to capture a near-complete 360$^\circ$ view comprising both the object and environment. RGB images, (raw) depth maps, and inertial streams are recorded; inertials are synchronized with RGB images and depth maps using nearest-neighbor timestamp matching.

Given the captured RGB images and raw depth maps, we estimate camera poses using ORB-SLAM3 \cite{campos2021orb} RGB-D mode. This yields a set of camera-to-world poses $\{ g_{\text{world} \leftarrow \text{cam}}^{(t)} \}_{t=1}^{T}$ for $T$ frames. We manually inspect the estimated trajectories and excluded or reprocessed sequences with tracking failures. Trajectories are used in all subsequent stages, e.g., point cloud construction, object registration and insertion, and ground truth generation. Note: inertials from the Intel RealSense D455 are recorded, synchronized to RGB images and depth maps, and released, but are not used for pose estimation in our processing pipeline.

As is common for hand-held, commodity depth sensors, the raw depth maps are noisy and are missing measurements near object boundaries. Depth maps tend to miss thin structures; (semi-)reflective surfaces and challenging lighting conditions also cause erroneous depth values. When aggregated as point clouds, raw depth maps lead to noisy, incomplete, and distorted 3D reconstructions of both the target objects and the surrounding (background) environment.

\textbf{Generating Ground-truth Depth Maps.} 
In order to obtain ground-truth 3D reconstructions of thin objects, we propose a data processing pipeline that (1) obtain a high-fidelity reconstruction (point cloud) of the thin object via a structured-light 3D scanner, (2) register the high-fidelity reconstruction to the raw noisy and incomplete point cloud of the object, (3) insert high-fidelity reconstruction into the raw point cloud (effectively replacing the noisy object) to yield a refined point cloud, and (4) re-project the refined point cloud onto the image plane corresponding to each camera pose to obtain ground-truth depth maps. Note: there are multiple filtering steps to reject outliers and clean the raw point cloud, in addition to insertion, to yield the refined point cloud. 

\begin{figure}[t]
    \centering
   \includegraphics[width=0.99\linewidth]{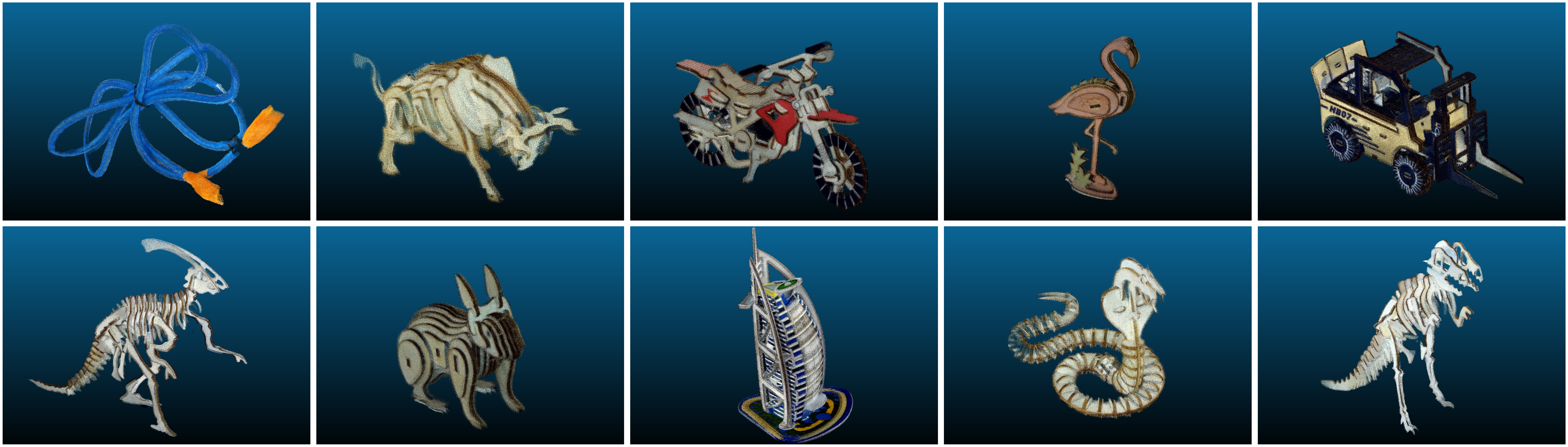}
   \vspace{-2mm}
   \caption{\textit{3D scans} of representative thin objects from our real-world dataset.}
   \label{fig:3d_scans}
   \vspace{-1mm}
\end{figure}

\textit{(1) Thin object 3D scanning.} 
To capture the fine structural details of the thin object, we use the 3DMakerPro Moose, a high-precision single-frame structured-light 3D scanner, to scan each object. The object is placed on a turntable for the capture. The scanner produces a high-fidelity, dense point cloud (see Fig. \ref{fig:teaser} and \ref{fig:3d_scans}). Any missing regions are identified by manual inspection; the object is further scanned and the resulting point cloud is merged and cleaned by manual removal of outliers. We denote the final point cloud as $P_{\text{scan}}$.

\textit{(2) Point cloud registration.} 
To register $P_{\text{scan}}$ to the aggregated raw point cloud robustly, we first isolate the subset of points corresponding to the object of interest in the raw point cloud; we then compute the transformation between the  scanned object to the object. To do so, we segment the object of interest using  SAM2~\cite{ravi2024sam} in each of the RGB images, which are spatially aligned with the raw depth maps. This yields a set of binary masks $\{M^{(t)}\}_{t=1}^{T}$, where $M^{(t)} \in \{0,1\}^{H \times W}$ denotes the object mask for frame at time $t$ with height $H$ and width $W$. To reduce artifacts while retaining sufficient coverage of the 3D scene, we uniformly sample frames from each sequence and discard frames for which more than 50\% of all depth pixels are invalid. For each depth map $z^{(t)}$ at $t$, we select the depth pixels within $M^{(t)}$, backproject them into 3D using the camera intrinsics $K \in \mathbb{R}^{3 \times 3}$, and transform the resulting points into world coordinates using the estimated camera-to-world pose. Aggregating these masked points across frames yields a raw point cloud of the object, denoted $P_{\text{obj}}$, which serves as the registration target for aligning $P_{\text{scan}}$ to the captured 3D scene.
\begin{equation}
    P_{\text{obj}} = \bigcup_{t=1}^{T} g_{\text{world} \leftarrow \text{cam}}^{(t)} \, K^{-1} \, \bar{x} \, z^{(t)}(x),
\end{equation}
where $x \in \{ u \in \Omega \mid M^{(t)}(u) = 1 \}$ denotes the image coordinates of the object, $\bar{x}$ their homogeneous coordinates, and $\Omega$ the image space. Since $P_{\text{obj}}$ contains floating artifacts, we apply statistical outlier removal \cite{rusu20113d} before registration. See Supp. Mat. for more details. 

Given that $P_{\text{scan}}$ and $P_{\text{obj}}$ are both in metric scale, we perform a two-stage rigid registration to transform $P_{\text{scan}}$ to world coordinates via $g_{\text{world} \leftarrow \text{scan}}$. To determine correspondence between the two point clouds, we manually select three or more pairs of corresponding points between $P_{\text{scan}}$ and $P_{\text{obj}}$. These sparse correspondences are used to estimate a rigid transformation, which provides an initial coarse alignment for subsequent refinement. Second, we utilize this as an initialization for Iterative Closest Point (ICP) algorithm, which iteratively establishes closest-point correspondences and updates the rigid transform to minimize point-to-point distances between $P_{\text{scan}}$ and $P_{\text{obj}}$, yielding $g_{\text{world} \leftarrow \text{scan}} \in SE(3)$.

\begin{figure*}[t]
    \centering
   \includegraphics[width=\linewidth]{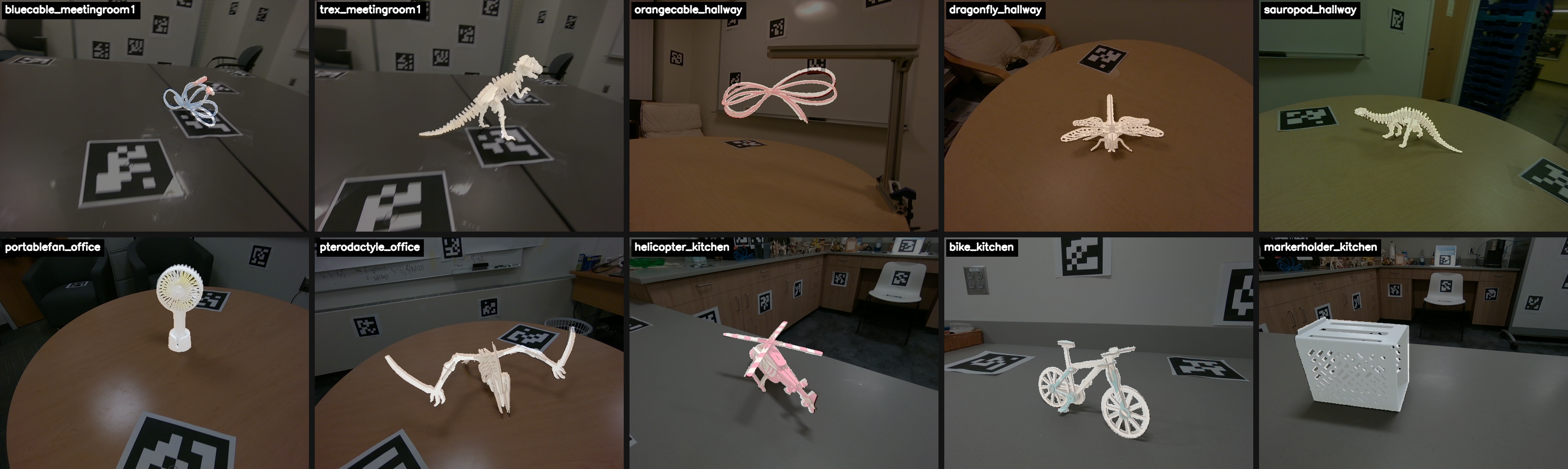}
   \vspace{-6mm}
   \caption{\textit{Segmentation masks.} Reprojected point clouds are overlaid as masks onto RGB images. This demonstrates the result of our highly accurate point cloud registration.}
   \label{fig:segmentation_overlay}
   \vspace{-0mm}
\end{figure*}

\textit{(3) Object insertion.} 
Given the set of backprojected points $P_{\text{raw}}$ from the raw depth maps, we remove the points corresponding to the object $P_{\text{obj}}$ and insert the scanned object. This yields the final refined point cloud: 
\begin{equation}
P = \left( P_{\text{raw}} \setminus P_{\text{obj}} \right)
\cup
\left\{ g_{\text{world} \leftarrow \text{scan}} \, p \;\middle|\; p \in P_{\text{scan}} \right\}.
\label{eqn:object_insertion}
\end{equation}

\begin{figure*}[t]
    \centering
    \includegraphics[width=\linewidth]{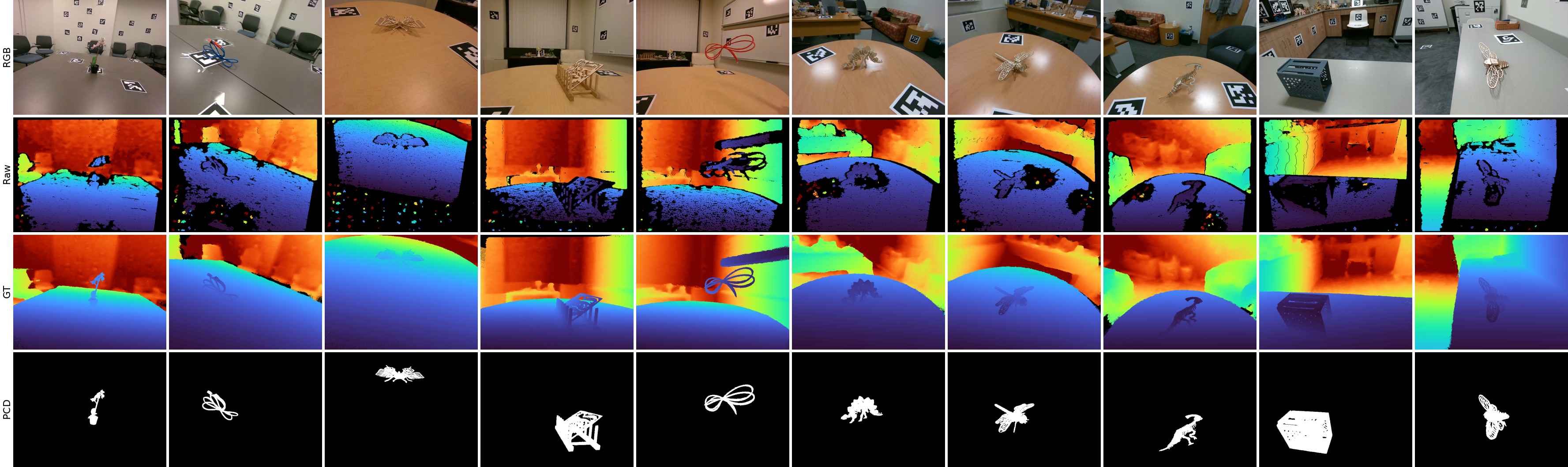}
    \vspace{-6mm}
    \caption{\textit{Ground-truth depth map and object segmentation masks}. Each object is scanned and inserted into an aggregated and refined point cloud to yield ground truth.}
    \label{fig:ground_truth_depth_segmentation}
    \vspace{-1mm}
\end{figure*}

\textit{(4) Reprojection of refined point cloud as ground truth.} 
For each frame at $t$, we project $P$ as depth map using the camera intrinsics $K$ and the inverse of the camera-to-world pose, $g_{\text{cam} \leftarrow \text{world}}^{(t)}$, and rasterize it as ground truth $\hat{D}^{(t)} \in \mathbb{R}_+^{H \times W}$: 
\begin{equation}
    \hat{D}^{(t)} = \mathcal{R} \left(K \ g_{\text{cam} \leftarrow \text{world}}^{(t)}  P\right),
\end{equation}
where $\mathcal{R}$ denotes canonical perspective projection and rasterization. This also produces a thin object projection mask $\hat{M}^{(t)} \in \{0 , 1\}^{H \times W}$, which also serves as a high-quality ground-truth object segmentation (see Fig \ref{fig:segmentation_overlay}). Note that we use two treatments of points inside and outside the object mask: Inside $\hat{M}^{(t)}$, we apply a standard $z$-buffer and retain the nearest projected depth at each pixel, which preserves thin structures and prevents background points from overwriting the object; outside $\hat{M}^{(t)}$, we use a $k$-median buffer, where for each pixel, we collect the closest $k$ projected depth values and assign their median value. This suppresses floating artifacts and produces smoother background surfaces and object boundaries. See Fig. \ref{fig:ground_truth_depth_segmentation} for representative examples.

\textbf{Tactile Data Collection.}
We collect vision-based tactile data using a DIGIT tactile sensor. For each object, we manually select eight contact locations that cover representative local surface properties, including edges, corners, and distinctive textures. We first annotate these contact locations on the registered high-fidelity 3D scan $P_{\text{scan}}$, which are transformed to 3D coordinates in the world coordinate frame via Eq. \ref{eqn:object_insertion}. At each selected location, the DIGIT sensor is manually pressed against the object surface while recording a short tactile video of $\approx$200 frames. Each recording captures light- to full-forced press contact state represented as 3-channel images, and reflects the local surface deformations.

\textbf{Semantic Segmentation Annotations.}
We annotate each sequence for semantic segmentation using CVAT, an online data annotation tool. For all scenes, we predefine a set of labels, e.g., furniture (shelf, table), room structures (wall, floor), and common indoor/outdoor objects (bench, printer, whiteboard). Annotators upload the captured RGB image sequence and manually annotate key frames using the built-in Segment Anything interactor. The masks are then propagated through the sequence using the SAM2 tracker and manually corrected for tracking failures, before being exported as semantic segmentation annotations.

\section{Dataset, Benchmark, and a Baseline}
VisTa3D consists of synchronized RGB images, depth maps, and tactile response maps, each associated with inertial measurements, camera poses, camera calibration, ground-truth depth maps, and segmentation masks (Fig. \ref{fig:teaser}). Each high-fidelity 3D scanned object is available as point cloud and mesh forms. While VisTa3D aims to benchmark thin object 3D reconstruction, it can also support segmentation and localization. It consists of synthetic and real-world components with the benchmark being a subset of the real-world component; the remaining data are available for training and pretraining (e.g., on synthetic data).

\begin{figure*}[t]
    \centering
    \includegraphics[width=\linewidth]{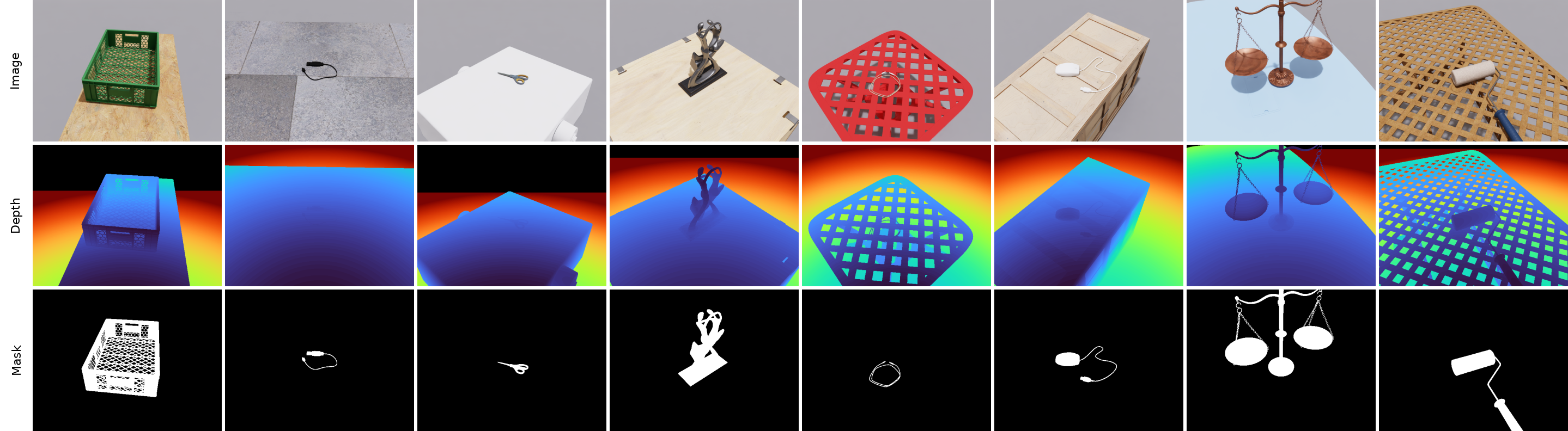}
    \vspace{-5mm}
    \caption{\textit{The synthetic dataset.} Examples of RGB images, depth maps, and object masks.}
    \label{fig:syn-dataset}
    \vspace{-2mm}
\end{figure*}

\textbf{Synthetic Data.}
The synthetic component is constructed in NVIDIA Isaac Sim 5.0. Each scene is created by arranging 18 thin objects on 1 of 9 tabletop configurations, resulting in a total of 162 scenes. For each scene, we provide RGB images, depth maps, camera intrinsics, camera poses, IMU measurements, object masks, tactile response maps from eight manually selected tactile contact locations, and object meshes (see Fig. \ref{fig:syn-dataset}). Each scene contains 500 frames covering a 360-degree view of the tabletop setup. To avoid overly regular turntable trajectories, we randomize camera elevation and distance, producing diverse viewpoints while maintaining object visibility. Objects comprise of wires, cables, figurines, and common household objects (scissors, paint roller, basket, corner shelf), and are modeled as rigid. The virtual camera was calibrated to match the Intel RealSense D455 used in the collection of real-world data.

\textbf{Real-World Data.}
We collected 52 thin objects in 8 different environments, including 6 indoor (meeting room, hallway, office, kitchen, cubicle, lobby) and 2 outdoor (bench, staircase) locations with varying background, lighting conditions, and motion trajectories -- resulting in 225 scenes in total. Our objects include wires, cables, wooden figurines (e.g., dinosaurs, insects, buildings, vehicles), plastic plants, and common office and household objects (e.g., wireless charger, portable fan, phone stand, dish rack). For each object, we provide a high-fidelity 3D scan (see Fig. \ref{fig:3d_scans}) using 3DMakerPro Moose, a high-precision structured-light scanner with 0.03 mm scanning accuracy, 0.07 mm resolution, and 15--1500 mm working range. Each point cloud contains 5--10M valid points. Each sequence contains 300--1000 frames covering $\approx$360 degrees around the object and surrounding environment. 
The Intel RealSense RGB image and depth map streams are synchronized at 10 Hz with 480 $\times$ 640 resolution; note that the effective range of the depth values is between $\approx$0.2 m and $\approx$5 m. Inertial streams comprise of gyroscope (angular velocity) and accelerometer (linear acceleration), each represented as a vector in $\mathbb{R}^3$, recorded at 200~Hz. They are synchronized to the image frames using nearest-neighbor timestamp matching. 

\textbf{The Benchmark.}
The VisTa3D benchmark focuses on 3D reconstruction of thin objects and covers multiple reconstruction paradigms, including MDE, MVS, MDC and NVS. Only our real data component is used for benchmarking, while the synthetic component is used for pretraining. We benchmark the baseline methods based on two tasks: depth estimation as the primary and novel view synthesis as the secondary task. For depth estimation, we split the data at the sequence level, randomly selecting 37 sequences from the full set of 225 sequences for testing and using the remaining sequences for training. This results in 88698 training frames and 13204 testing frames in total. For novel view synthesis, we instead split frames within each sequence by uniformly selecting one out of every five frames for training and using the remaining frames for testing.

\textbf{Evaluation Protocol.}
We evaluate the predicted depth maps using standard metrics against ground truth: A1--A3 (with thresholds 1.05, $1.05^2$, $1.05^3$), AbsRel, MAE, logMAE, logRMSE, and RMSE. We evaluate depth only within the range from 0.2 m to 5.0 m, corresponding to the reliable sensing range of the Intel RealSense D455. 
For NVS methods, we also report PSNR, SSIM, and LPIPS for rendered RGB images. See Supp. Mat. for definitions of all metrics. 

As depth estimators can be scale-less or metric-scale, we evaluate each under 3 protocols to mitigate confounding factors brought by scale. \textit{Default} reports metrics on the raw model outputs. \textit{Median Scale} rescales outputs per frame using a median-based scale factor and \textit{Linear Fit} applies a per-frame scale and shift -- both estimated from the ground truth. To focus on thin structures, we consider 2 evaluation modes: \textit{general} (without masking) and \textit{thin-object focused} (masked). In the \textit{general} evaluation, we compute errors over all pixels with valid ground truth. In thin-object focused evaluation, we compute errors only within the binary mask $\hat{M}^{(t)}$ of the projected object point cloud $P_\text{scan}$.

\textit{Summary of benchmark.}
We evaluate 11 representative methods spanning MDE, MVS, MDC, and NVS. We found that all methods struggle in \textit{thin-object focused} evaluation. In response, we hypothesize that tactile data can provide the necessary local deformation information to aid thin object reconstruction. To this end, we propose the first 3D reconstruction method using visual, range, and tactile inputs as a baseline, bringing the total to 12 methods.

\textbf{Tactile-DC: A Visual-Range-Tactile Baseline.}
\label{sec:tactiledc}
To study whether localized tactile observations can aid in their thin object reconstruction, we introduce \textit{Tactile-DC}, inspired by existing works in MDC \cite{li2020multi,park2020non,wong2021unsupervised}. To the best of our knowledge, this is the first baseline to incorporate RGB image, sparse depth map, and tactile responses for thin-object depth completion. By exploiting tactile cues and local connectivity of surfaces, Tactile-DC aims to recover fine structures in contact regions that are difficult to resolve from images, point clouds, or both.

\begin{figure}[t]
    \centering
    \includegraphics[width=0.84\linewidth]{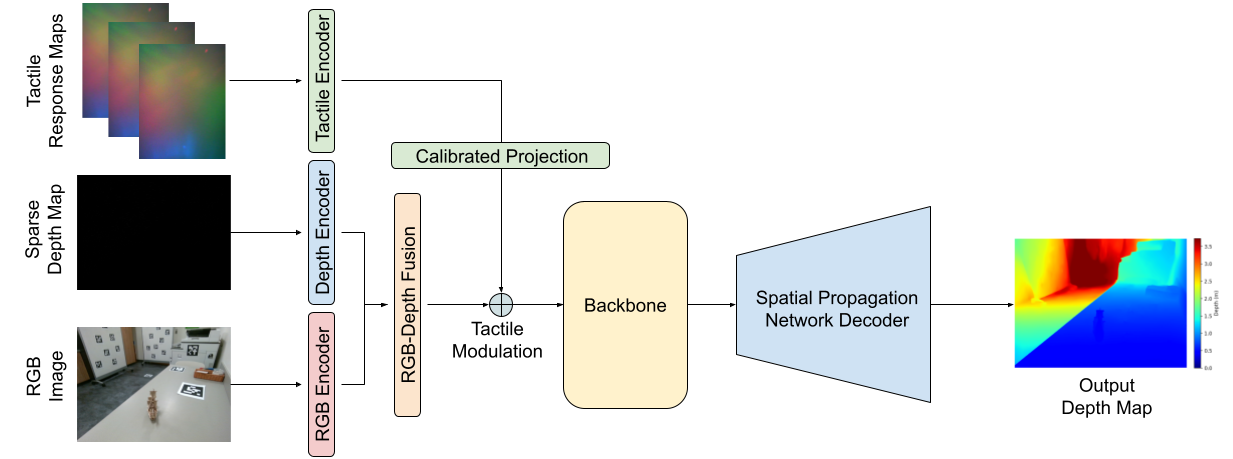}
    \vspace{-2mm}
    \caption{\textit{Tactile-DC.} Each input is encoded separately. RGB and sparse depth feature maps are first fused together. Tactile responses are encoded and projected onto the image plane based on contact coordinates, camera intrinsic and extrinsic calibration.}
    \label{fig:Tactile-DC_architecture}
    \vspace{-2mm}
\end{figure}

Tactile-DC follows a conventional depth completion architecture and employs shallow RGB image and sparse depth map encoders. The RGB image and sparse depth latent feature maps are fused together by a convolutional layer. We additionally introduce a tactile encoder that encodes each tactile response map separately. As each tactile response map is associated with a 3D contact location in the world coordinate frame, we use camera intrinsics and poses to project 3D contact locations onto the image plane, following a calibrated projection ~\cite{wong2021unsupervised}. 

Each tactile response map is encoded as feature map of size $D \times h \times w$, where $D$ is its latent dimension and $h$ and $w$ are its spatial dimensions. We then apply a $1 \times 1$ convolutional layer with batch normalization and ReLU followed by a global average pooling to compress each feature map to a $D \times 1 \times 1$ latent vector. To inject tactile information into the RGB-depth representation, each tactile latent vector is added to the locations of their projection on the image plane to modulate the fused feature maps of the RGB image and sparse depth map. Note: contact locations accounts for feature map resolution. The modulated feature map is then fed to a backbone that follows \cite{zhang2023completionformer}, and a Spatial Propagation Network \cite{park2020non} decodes it into a depth map. See Fig.~\ref{fig:Tactile-DC_architecture} for the architecture.

\begin{table*}[h!]
\centering
\caption{Evaluation metrics across models and protocols on the test set.}
\vspace{-2mm}
\label{tab:depth_results}
\resizebox{0.84\textwidth}{!}{
\begin{tabular}{l l r r r r r r r r} 
\toprule
\textbf{Model} & \textbf{Protocol} & \textbf{A1} & \textbf{A2} & \textbf{A3} & \textbf{AbsRel} & \textbf{logMAE} & \textbf{logRMSE} & \textbf{MAE} & \textbf{RMSE} \\
\midrule

\rowcolor{gray!15}
\multicolumn{10}{c}{\textbf{General Evaluation}} \\
\midrule

\multirow{3}{*}{UniDepth V2} 
  & Default & 0.002 & 0.004 & 0.007 & 7.981 & 0.782 & 1.890 & 4.651 & 4.750 \\
  & Linear Fit & 0.201 & 0.323 & 0.420 & 0.354 & 0.129 & 0.362 & 0.299 & 0.392 \\
  & Median Scale & 0.152 & 0.265 & 0.359 & 0.395 & 0.164 & 0.477 & 0.420 & 0.582 \\
\midrule

\multirow{3}{*}{UniDepth V1} 
  & Default & 0.007 & 0.014 & 0.022 & 7.047 & 0.721 & 1.780 & 3.898 & 4.032 \\
  & Linear Fit & 0.139 & 0.234 & 0.302 & 0.642 & 0.199 & 0.545 & 0.441 & 0.536 \\
  & Median Scale & 0.093 & 0.169 & 0.235 & 0.588 & 0.232 & 0.652 & 0.544 & 0.716 \\
\midrule

\multirow{3}{*}{DepthAnything V2} 
  & Default & 0.002 & 0.005 & 0.007 & 4.201 & 0.657 & 1.563 & 3.700 & 4.360 \\
  & Linear Fit & 0.097 & 0.191 & 0.278 & 0.488 & 0.166 & 0.450 & 0.387 & 0.479 \\
  & Median Scale & 0.061 & 0.119 & 0.173 & 1.214 & 0.315 & 0.898 & 1.566 & 2.552 \\
\midrule

\multirow{3}{*}{DepthAnything V1} 
  & Default & 0.006 & 0.012 & 0.018 & 4.241 & 0.652 & 1.565 & 3.612 & 4.281 \\
  & Linear Fit & 0.073 & 0.145 & 0.215 & 0.633 & 0.202 & 0.535 & 0.467 & 0.556 \\
  & Median Scale & 0.054 & 0.106 & 0.156 & 1.093 & 0.303 & 0.856 & 1.314 & 2.189 \\
\midrule

\multirow{3}{*}{VGGT} 
  & Default & 0.060 & 0.117 & 0.172 & 0.486 & 0.215 & 0.545 & 0.496 & 0.633 \\
  & Linear Fit & 0.229 & 0.409 & 0.550 & 0.202 & 0.086 & 0.279 & 0.175 & 0.261 \\
  & Median Scale & 0.231 & 0.401 & 0.521 & 0.215 & 0.096 & 0.307 & 0.237 & 0.359 \\
\midrule

\multirow{3}{*}{MapAnything} 
  & Default & 0.511 & 0.627 & 0.676 & 0.307 & 2.096 & 10.965 & 0.210 & 0.495 \\
  & Linear Fit & 0.242 & 0.376 & 0.452 & 0.393 & 0.137 & 0.447 & 0.254 & 0.395 \\
  & Median Scale & 0.187 & 0.303 & 0.378 & 0.521 & 2.152 & 10.967 & 0.565 & 0.833 \\
\midrule

\multirow{3}{*}{OmniDC} 
  & Default & 0.834 & 0.881 & 0.897 & 0.409 & 0.054 & 0.312 & 0.139 & 0.460 \\
  & Linear Fit & 0.484 & 0.569 & 0.624 & 0.367 & 0.102 & 0.320 & 0.212 & 0.323 \\
  & Median Scale & 0.579 & 0.682 & 0.730 & 0.297 & 0.090 & 0.318 & 0.201 & 0.418 \\
\midrule

\multirow{3}{*}{Tactile-DC} 
  & Default & \textbf{0.844} & \textbf{0.900} & \textbf{0.920} & \textbf{0.211} & \textbf{0.041} & \textbf{0.247} & \textbf{0.091} & \textbf{0.347} \\
  & Linear Fit & \textbf{0.635} & \textbf{0.744} & \textbf{0.810} & \textbf{0.201} & \textbf{0.066} & \textbf{0.262} & \textbf{0.112} & \textbf{0.215} \\
  & Median Scale & \textbf{0.662} & \textbf{0.774} & \textbf{0.821} & \textbf{0.200} & \textbf{0.054} & \textbf{0.244} & \textbf{0.133} & \textbf{0.273} \\
  
\midrule

\rowcolor{gray!15}
\multicolumn{10}{c}{\textbf{Object Evaluation}} \\
\midrule

\multirow{3}{*}{UniDepth V2} 
  & Default & 0.000 & 0.000 & 0.000 & 13.108 & 1.041 & 2.402 & 4.831 & 4.848 \\
  & Linear Fit & 0.092 & 0.170 & 0.235 & 0.760 & 0.211 & 0.515 & 0.287 & 0.313 \\
  & Median Scale & 0.080 & 0.162 & 0.243 & 0.695 & 0.195 & 0.466 & 0.273 & 0.290 \\
\midrule

\multirow{3}{*}{UniDepth V1} 
  & Default & 0.000 & 0.000 & 0.000 & 12.498 & 1.014 & 2.339 & 4.582 & 4.597 \\
  & Linear Fit & 0.049 & 0.092 & 0.127 & 1.432 & 0.341 & 0.814 & 0.540 & 0.566 \\
  & Median Scale & 0.033 & 0.067 & 0.103 & 1.255 & 0.301 & 0.710 & 0.485 & 0.501 \\
\midrule

\multirow{3}{*}{DepthAnything V2} 
  & Default & 0.000 & 0.000 & 0.000 & 4.273 & 0.673 & 1.558 & 1.646 & 1.714 \\
  & Linear Fit & 0.058 & 0.115 & 0.170 & 0.903 & 0.244 & 0.573 & 0.355 & 0.366 \\
  & Median Scale & 0.087 & 0.169 & 0.242 & 0.465 & 0.233 & 0.564 & 0.192 & 0.249 \\
\midrule

\multirow{3}{*}{DepthAnything V1} 
  & Default & 0.000 & 0.000 & 0.000 & 4.602 & 0.697 & 1.613 & 1.781 & 1.857 \\
  & Linear Fit & 0.025 & 0.053 & 0.086 & 1.161 & 0.298 & 0.696 & 0.448 & 0.454 \\
  & Median Scale & 0.071 & 0.140 & 0.205 & 0.554 & 0.237 & 0.574 & 0.227 & 0.281 \\
\midrule

\multirow{3}{*}{VGGT} 
  & Default & 0.064 & 0.125 & 0.186 & 0.563 & 0.205 & 0.484 & 0.214 & 0.228 \\
  & Linear Fit & 0.119 & 0.235 & 0.360 & 0.321 & 0.104 & 0.297 & 0.141 & 0.230 \\
  & Median Scale & 0.221 & 0.386 & 0.510 & 0.262 & 0.091 & 0.277 & 0.108 & 0.186 \\
\midrule

\multirow{3}{*}{MapAnything} 
  & Default & 0.165 & 0.288 & 0.375 & 0.485 & 3.307 & 13.607 & 0.195 & 0.293 \\
  & Linear Fit & 0.035 & 0.072 & 0.113 & 0.668 & 0.207 & 0.520 & 0.262 & 0.302 \\
  & Median Scale & 0.144 & 0.249 & 0.324 & 0.592 & 3.318 & 13.607 & 0.241 & 0.354 \\
\midrule

\multirow{3}{*}{OmniDC} 
  & Default & 0.687 & 0.772 & 0.817 & 0.274 & 0.058 & 0.260 & 0.101 & 0.238 \\
  & Linear Fit & 0.342 & 0.423 & 0.472 & 0.513 & 0.142 & 0.390 & 0.194 & 0.263 \\
  & Median Scale & 0.439 & 0.589 & 0.659 & 0.274 & 0.102 & 0.311 & 0.106 & 0.192 \\
\midrule

\multirow{3}{*}{Tactile-DC} 
  & Default & \textbf{0.719} & \textbf{0.811} & \textbf{0.852} & \textbf{0.205} & \textbf{0.047} & \textbf{0.226} & \textbf{0.078} & \textbf{0.205} \\
  & Linear Fit & \textbf{0.487} & \textbf{0.585} & \textbf{0.645} & \textbf{0.312} & \textbf{0.090} & \textbf{0.288} & \textbf{0.122} & \textbf{0.214} \\
  & Median Scale & \textbf{0.540} & \textbf{0.676} & \textbf{0.750} & \textbf{0.220} & \textbf{0.070} & \textbf{0.249} & \textbf{0.086} & \textbf{0.166} \\

\bottomrule
\end{tabular}}
\vspace{-3mm}
\end{table*}

\begin{figure}[h!]
    \centering
    \includegraphics[width=0.88\linewidth]{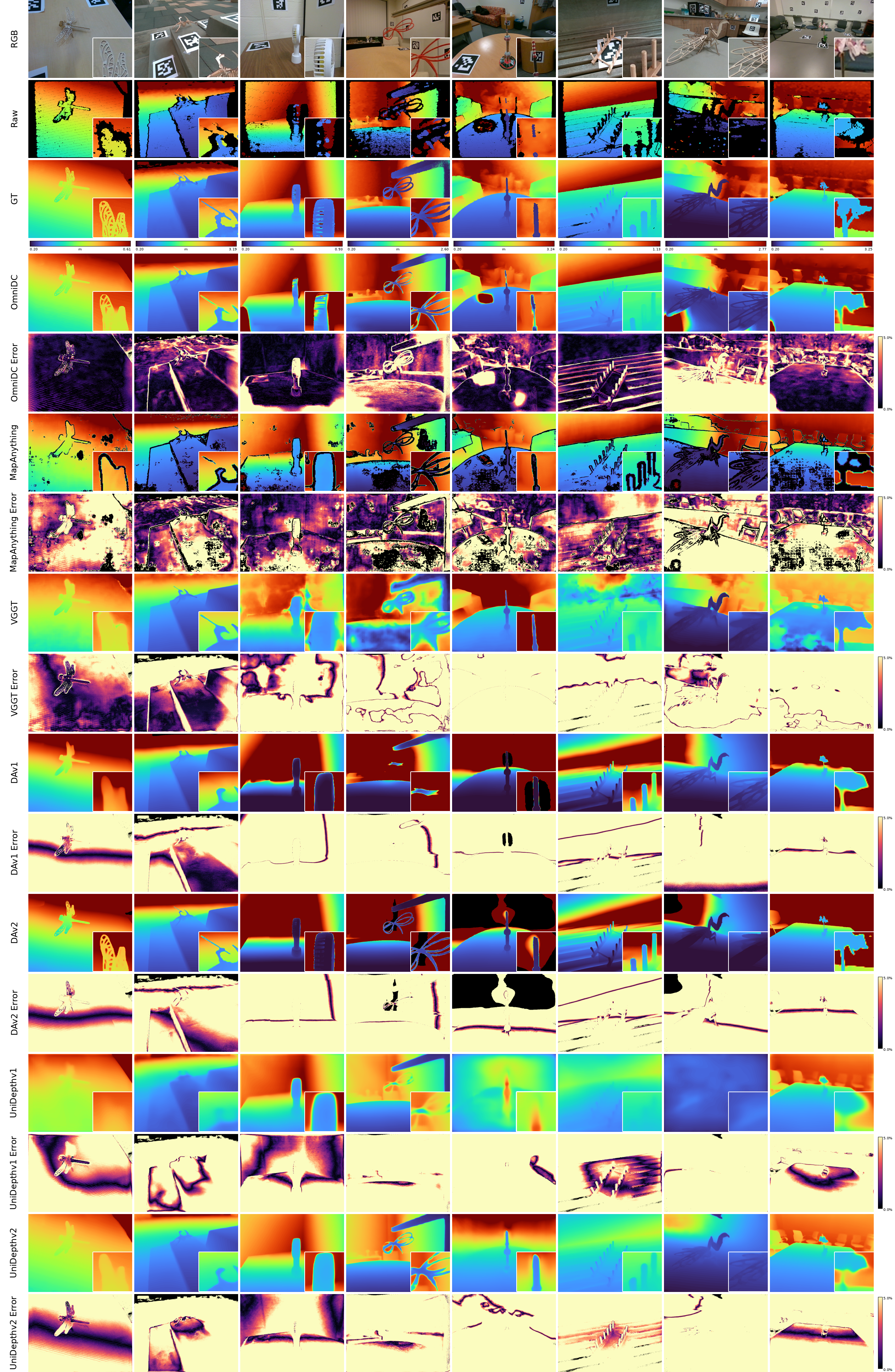}
    \vspace{-2mm}
    \caption{
    \textit{Results on VisTa3D.} Methods struggle on thin objects. Best viewed in 5$\times$.
    }
    \label{fig:depth-error}
    \vspace{-2mm}
\end{figure}

\begin{figure*}[h!]
    \centering
    \includegraphics[width=0.88\linewidth]{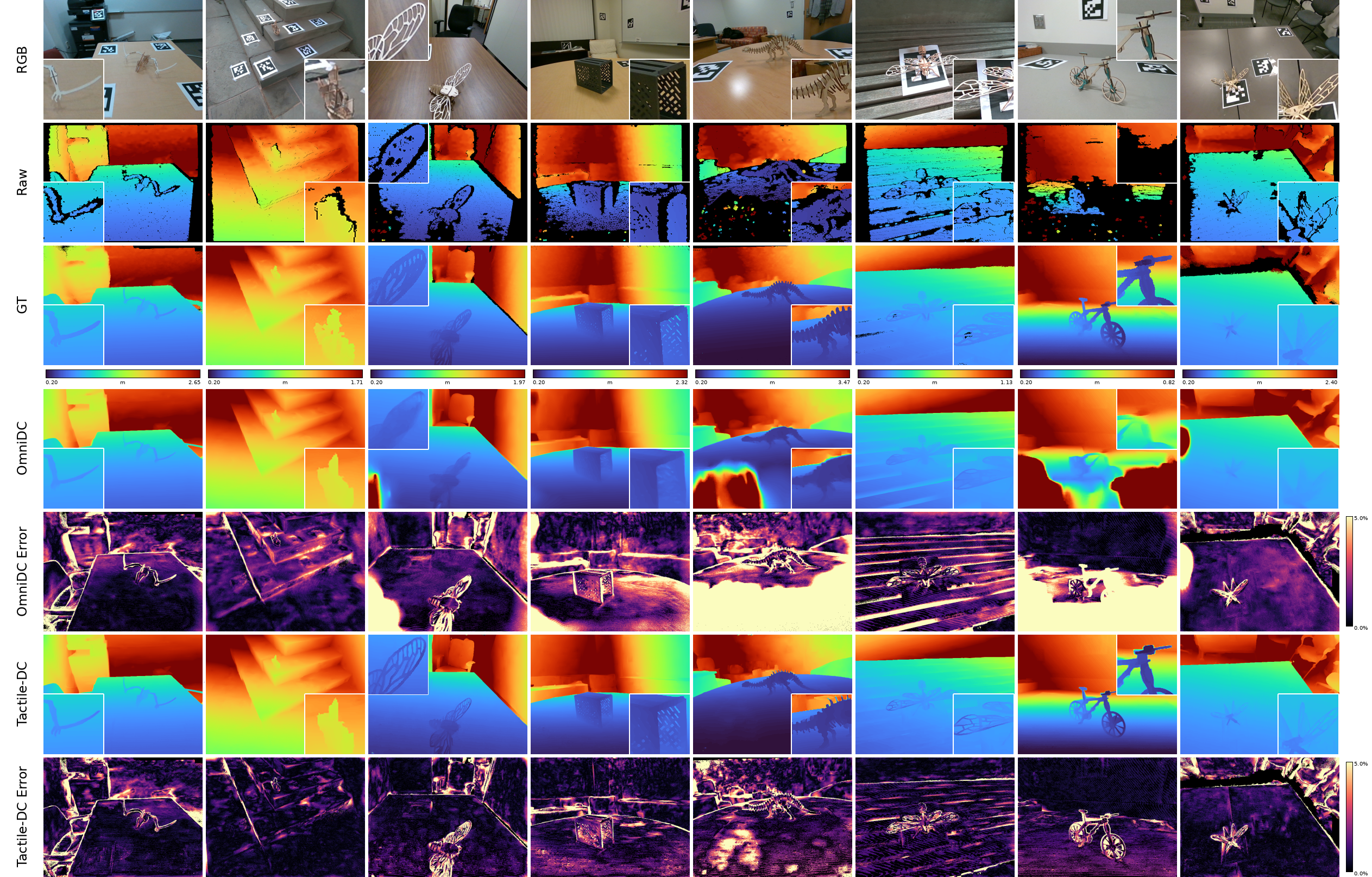}
    \vspace{-2mm}
    \caption{\textit{Comparison between OmniDC and our Tactile-DC.} Tactile-DC produces more accurate depth estimates, especially around thin-object areas.}
    \label{fig:depth-error-tactile}
    \vspace{-2mm}
\end{figure*}

\section{Experiments}
Implementation details, including preprocessing, model configurations, and training settings, are provided in Supp. Mat., along with ablations and analysis.

\textbf{Baselines.}
We benchmark 12 representative methods spanning NVS (2D \cite{huang20242d} and 3D Gaussian Splatting \cite{kerbl20233d}, Nerfacto \cite{tancik2023nerfstudio}, Depth-Nerfacto \cite{tancik2023nerfstudio}), MDE (DepthAnything V1 \cite{yang2024depthanythingv1}, V2 \cite{yang2024depthanythingv2}, UniDepth V1 \cite{piccinelli2024unidepth}, V2 \cite{piccinelli2025unidepthv2}), MVS (VGGT \cite{wang2025vggt}, MapAnything \cite{keetha2026mapanything}), MDC (OmniDC \cite{zuo2025omni}), and the proposed Tactile-DC. For all methods except Tactile-DC, we use open-source code and checkpoints when available, and modify only data paths and input format to match our protocol.

\textbf{Depth Estimation Results.}
Table ~\ref{tab:depth_results} shows that baselines struggle on VisTa3D, particularly on thin-object regions. Compared with general (without masking) evaluation, thin-object focused (masked) evaluation generally leads to lower A1--A3 scores, indicating that thin object pixels are harder to reconstruct than the full image. Note: errors do not always increase under the masked setting because the thin object covers a different and often smaller depth range, but the consistent decrease in A1--A3 shows that fewer thin-object pixels are reconstructed within acceptable thresholds. Fig. ~\ref{fig:depth-error} further illustrates this: error is higher near thin structures and object boundaries. This supports our claim that thin structures remain challenging for current reconstruction methods. 

Among existing baselines, OmniDC performs best overall, especially under the default protocol. This is expected as OmniDC is a depth completion method and directly uses sparse metric depth maps as input. Qualitatively, OmniDC produces depth maps that are globally close to the ground truth, but still show considerable errors around thin structures and object boundaries. Linear fitting or median scaling does not consistently improve OmniDC as its main failure mode is not global scale mismatch but the recovery of thin structures.

In contrast, MDE models such as UniDepth V1/V2, DepthAnything V1/V2 are strongly affected by scale ambiguity. All four methods perform poorly under the default protocol, especially in thin-object focused evaluation, where A1--A3 are close to zero for several settings. Linear fitting and median scaling substantially improve their results, showing that part of the error comes from global scale mismatch. However, even after scale alignment, their accuracies and performance for the object region remain low in the qualitative results, suggesting that another error mode exists in recovering thin structures. Between model versions, UniDepth V2 generally improves over V1, and DepthAnything V2, improves over V1 after scale correction, but both still remain far below MDC (e.g., A1 below 0.10 for UniDepth V1/V2, and below 0.09 for DepthAnything V1/V2). 

For MVS, VGGT performs better than MDEs, showing the benefit of multiple views; its qualitative results preserve more scene structure than MDE and exhibit smoother, more coherent outputs. However, VGGT still misses or over-smooths thin details, and its thin-object focused performance remains low. MapAnything shows a large gap between general and thin-object focused evaluation despite using images, calibration, and sparse depth, which are sufficient to ground estimates to metric scale; it also often produces noisy or incomplete estimates with scattered invalid regions and errors around thin objects and foreground-background transitions. This indicates that MVS methods may not preserve small thin-object regions reliably, even when sparse metric depth is provided. 

\textbf{Tactile-DC} Among all methods, Tactile-DC achieves the best performance with A1/A2/A3 of 0.844/0.900/0.920 for general evaluation, and 0.719/0.811/ 0.852 for object evaluation. It consistently outperforms OmniDC, across all protocols under both general and thin-object focused evaluation (see Fig. ~\ref{fig:depth-error-tactile} for comparison). In particular, Tactile-DC obtains the best A1--A3 scores and the lowest errors in all settings, demonstrating that tactile data provide useful cues for recovering thin structures that are difficult to infer from RGB images and sparse depth maps. The advantage of Tactile-DC is especially clear under thin-object focused evaluation, where visual or sparse-depth cues are often weak.

\textit{Summary:} Results show a clear hierarchy amongst depth estimation methods: MDE struggle most with scale and feed-forward MVS improves global structure; both still miss thin object boundaries and details. MDC benefit from sparse metric cues, and Tactile-DC performs the best  by incorporating tactile information. 

\textbf{NVS Results.} We report PSNR, SSIM, and LPIPS in Table ~\ref{tab:nvs_results}. Nerfacto achieves the best overall performance, obtaining the highest PSNR and SSIM as well as the lowest LPIPS. Depth-Nerfacto performs slightly worse despite using depth supervision, suggesting that the raw sensor depth may not consistently benefit NVS, especially when depth sensing is noisy or incomplete around thin structures. Among Gaussian-splatting (GS) methods, 3DGS outperforms 2DGS across all three metrics, indicating better rendering quality. However, GS methods are worse than Nerfacto-based methods. These results suggest that current NVS methods can synthesize plausible novel views, but sparse input views and occlusions remain challenging. 
Due to space limitations, we defer the NVS depth estimation results and ablations of tactile inputs to Supp. Mat.

\begin{table}[t!]
\centering
\scriptsize
\caption{Novel view synthesis performance on the test set.}
\label{tab:nvs_results}
\vspace{-2mm}
\begin{tabular}{lccc}
\toprule
\textbf{Method} & \textbf{PSNR} $\uparrow$ & \textbf{SSIM} $\uparrow$ & \textbf{LPIPS} $\downarrow$ \\
\midrule
2D Gaussian Splatting & 21.74 & 0.802 & 0.367 \\
3D Gaussian Splatting & 22.97 & 0.805 & 0.322 \\
Nerfacto              & 24.16 & 0.855 & 0.177 \\
Depth-Nerfacto        & 23.52 & 0.835 & 0.178 \\
\bottomrule
\end{tabular}
\vspace{-0mm}
\end{table}

\section{Discussion and Limitations}

While our dataset provides a comprehensive benchmark for thin-object reconstruction across diverse indoor and outdoor scenes, several limitations remain. First, the selection of real-world objects is constrained by the sensing characteristics of the Intel RealSense and the 3DMakerPro Moose, both of which perform unreliably on reflective, specular, or semi-transparent surfaces. Therefore, such objects are not included in our dataset. Second, despite covering a wide range of object categories, the scale of the dataset remains moderate, and real-world thin objects span a broader range of materials, thicknesses, and shapes than those represented here. Third, we assume the 3D scene is static and does not consider moving or deformable objects. Finally, MDE and MVS may benefit from task-specific finetuning, which we do not explore in this benchmark. Future work will expand object diversity, introduce controlled dynamic scenes for tasks like motion estimation \cite{lao2018extending,xiao2026triangular} and 4D reconstruction \cite{zhang2025monst3r}, and incorporate additional modalities to better capture failure modes of existing reconstruction pipelines.

\paragraph{\textbf{Acknowledgements.}}
This work is supported by NSF-2112562 Athena AI Institute and the Global Industrial Technology Cooperation Center (GITCC) through a grant agreement with the Korea Institute for Advancement of Technology (KIAT), project number P0028922.

\bibliographystyle{splncs04}
\bibliography{global,visionlab}

\newpage
\appendix

\begin{center}
{\Large{\textbf{
VisTa3D: A Dataset and Benchmark for \\
Thin Object Reconstruction from \\ 
\underline{Vis}ion, \underline{Ta}ctile, and \underline{3D} Point Clouds \\
\vspace{0.5cm}
SUPPLEMENTARY MATERIAL
}}}
\vspace{1cm}
\end{center}

\section{Additional Results}
We include here results that were unable to be presented in the main paper due to space limitations.

\textbf{Effect of Synthetic Data.}
As discussed in the main text, VisTa3D comprises real and synthetic components. We test whether our synthetic dataset of thin objects can be used to finetune existing methods to improve their performance. For this experiment, we choose the best competing method, OmniDC \cite{zuo2025omni}, restore from its checkpoint, and finetune it using synthetically rendered RGB images, sparse depth maps, and ground-truth depth maps.  Table ~\ref{tab:finetuning_omnidc} shows that, despite training on synthetic data, OmniDC still improved over the pretrained checkpoint released by the authors. To further validate the use of tactile data in 3D reconstruction of thin objects, we repeat this experiment with Tactile-DC, which follows the architecture of OmniDC, but augmented with a tactile encoder; we restore the checkpoint of OmniDC into network modules shared with Tactile-DC and finetune it on  RGB images, sparse depth maps, tactile response maps, and ground-truth depth maps. Table ~\ref{tab:finetuning_omnidc} validates that tactile data does aid in the 3D reconstruction of thin objects. Tactile-DC trained on synthetic data VisTa3D improves over both OmniDC released by the authors and OmniDC finetuned on the same synthetic data.

\textbf{Effect of Real Data.}
To test whether the synthetic can be useful for seeding the pretraining of 3D reconstruction models, we further finetune OmniDC and Tactile-DC, both pretrained on synthetic data, on the real component of VisTa3D. Table ~\ref{tab:finetuning_omnidc} shows consistent improvements for both OmniDC and Tactile-DC over their counterparts pretrained on the synthetic dataset. This demonstrates that indeed our synthetic dataset can be used for pretraining. Furthermore, our conclusions still hold: Tactile-DC trained on real data improves over OmniDC trained on the same. Below, we describe our results in detail.

\begin{table*}[t!]
\centering
\caption{Quantitative comparison between author-released OmniDC, fine-tuned OmniDC models, and Tactile-DC on the test set. Higher is better for A1--A3, while lower is better for all other metrics. The best result for each protocol is shown in bold.}
\label{tab:finetuning_omnidc}
\resizebox{0.99\textwidth}{!}{
\begin{tabular}{l l l r r r r r r r r}
\toprule
\textbf{Model} & \textbf{Training Data} & \textbf{Protocol} 
& \textbf{A1} & \textbf{A2} & \textbf{A3} 
& \textbf{AbsRel} & \textbf{logMAE} & \textbf{logRMSE} 
& \textbf{MAE} & \textbf{RMSE} \\
\midrule

\rowcolor{gray!15}
\multicolumn{11}{c}{\textbf{General Evaluation}} \\
\midrule

\multirow{9}{*}{OmniDC}
  & \multirow{3}{*}{Author-released}
  & Default      & 0.834 & 0.881 & 0.897 & 0.409 & 0.054 & 0.312 & 0.139 & 0.460 \\
  & 
  & Linear Fit   & 0.484 & 0.569 & 0.624 & 0.367 & 0.102 & 0.320 & 0.212 & 0.323 \\
  & 
  & Median Scale & 0.579 & 0.682 & 0.730 & 0.297 & 0.090 & 0.318 & 0.201 & 0.418 \\
\cmidrule(lr){2-11}

  & \multirow{3}{*}{Synthetic}
  & Default      & 0.819 & 0.881 & 0.903 & 0.278 & 0.058 & 0.310 & 0.107 & 0.387 \\
  & 
  & Linear Fit   & 0.526 & 0.625 & 0.680 & 0.255 & 0.076 & 0.311 & 0.149 & 0.258 \\
  & 
  & Median Scale & 0.623 & 0.741 & 0.793 & 0.237 & 0.057 & 0.321 & 0.151 & 0.340 \\
\cmidrule(lr){2-11}

  & \multirow{3}{*}{Synthetic + Real}
  & Default      & 0.834 & 0.888 & 0.910 & 0.247 & 0.048 & 0.273 & 0.107 & 0.355 \\
  & 
  & Linear Fit   & 0.627 & 0.726 & 0.780 & 0.220 & 0.069 & 0.303 & 0.119 & 0.226 \\
  & 
  & Median Scale & 0.631 & 0.744 & 0.795 & 0.220 & 0.056 & 0.275 & 0.147 & 0.284 \\
\midrule

\multirow{6}{*}{Tactile-DC}

  & \multirow{3}{*}{Synthetic}
  & Default      & 0.831 & 0.891 & 0.913 & 0.257 & 0.046 & 0.250 & 0.099 & 0.353 \\
  & 
  & Linear Fit   & 0.540 & 0.645 & 0.705 & 0.238 & 0.072 & 0.268 & 0.146 & 0.253 \\
  & 
  & Median Scale & 0.634 & 0.750 & 0.805 & 0.210 & 0.055 & 0.254 & 0.144 & 0.328 \\
\cmidrule(lr){2-11}

  & \multirow{3}{*}{Synthetic + Real}
  & Default      & \textbf{0.844} & \textbf{0.900} & \textbf{0.920} & \textbf{0.211} & \textbf{0.041} & \textbf{0.247} & \textbf{0.091} & \textbf{0.347} \\
  & 
  & Linear Fit   & \textbf{0.635} & \textbf{0.744} & \textbf{0.810} & \textbf{0.201} & \textbf{0.066} & \textbf{0.262} & \textbf{0.112} & \textbf{0.215} \\
  & 
  & Median Scale & \textbf{0.662} & \textbf{0.774} & \textbf{0.821} & \textbf{0.200} & \textbf{0.054} & \textbf{0.244} & \textbf{0.133} & \textbf{0.273} \\
\midrule

\rowcolor{gray!15}
\multicolumn{11}{c}{\textbf{Object Evaluation}} \\
\midrule

\multirow{9}{*}{OmniDC}
  & \multirow{3}{*}{Author-released}
  & Default      & 0.687 & 0.772 & 0.817 & 0.274 & 0.058 & 0.260 & 0.101 & 0.238 \\
  & 
  & Linear Fit   & 0.342 & 0.423 & 0.472 & 0.513 & 0.142 & 0.390 & 0.194 & 0.263 \\
  & 
  & Median Scale & 0.439 & 0.589 & 0.659 & 0.274 & 0.102 & 0.311 & 0.106 & 0.192 \\
\cmidrule(lr){2-11}

  & \multirow{3}{*}{Synthetic}
  & Default      & 0.702 & 0.794 & 0.838 & 0.230 & 0.051 & 0.238 & 0.086 & 0.214 \\
  & 
  & Linear Fit   & 0.379 & 0.480 & 0.538 & 0.404 & 0.116 & 0.341 & 0.152 & 0.234 \\
  & 
  & Median Scale & 0.475 & 0.642 & 0.719 & 0.235 & 0.073 & 0.266 & 0.091 & 0.186 \\
\cmidrule(lr){2-11}

  & \multirow{3}{*}{Synthetic + Real}
  & Default      & 0.717 & 0.807 & 0.848 & 0.218 & 0.049 & 0.232 & 0.082 & 0.213 \\
  & 
  & Linear Fit   & 0.474 & 0.578 & 0.638 & 0.347 & 0.100 & 0.297 & 0.131 & 0.223 \\
  & 
  & Median Scale & 0.527 & 0.645 & 0.720 & 0.234 & 0.073 & 0.261 & 0.090 & 0.167 \\
\midrule

\multirow{6}{*}{Tactile-DC}

  & \multirow{3}{*}{Synthetic}
  & Default      & 0.706 & 0.798 & 0.841 & 0.226 & 0.051 & 0.235 & 0.085 & 0.212 \\
  & 
  & Linear Fit   & 0.389 & 0.494 & 0.552 & 0.396 & 0.112 & 0.334 & 0.149 & 0.233 \\
  & 
  & Median Scale & 0.486 & 0.652 & 0.730 & 0.232 & 0.072 & 0.251 & 0.090 & 0.186 \\
\cmidrule(lr){2-11}

  & \multirow{3}{*}{Synthetic + Real}
  & Default      & \textbf{0.719} & \textbf{0.811} & \textbf{0.852} & \textbf{0.205} & \textbf{0.047} & \textbf{0.226} & \textbf{0.078} & \textbf{0.205} \\
  & 
  & Linear Fit   & \textbf{0.487} & \textbf{0.585} & \textbf{0.645} & \textbf{0.312} & \textbf{0.090} & \textbf{0.288} & \textbf{0.122} & \textbf{0.214} \\
  & 
  & Median Scale & \textbf{0.540} & \textbf{0.676} & \textbf{0.750} & \textbf{0.220} & \textbf{0.070} & \textbf{0.249} & \textbf{0.086} & \textbf{0.166} \\
\bottomrule

\end{tabular}}
\end{table*}

\textbf{Results.}
Table ~\ref{tab:finetuning_omnidc} shows that fine-tuning improves OmniDC in most settings, especially on thin-object regions. Compared with vanilla OmniDC, finetuning OmniDC with synthetic RGB and depth reduces default AbsRel from 0.274 to 0.230 and RMSE from 0.238 to 0.214 for object evaluation, showing that the synthetic component of VisTa3D already helps improve the model's performance on real thin-object depth completion. Tactile-DC finetuned with additional synthetic tactile data further improves over OmniDC finetuned on synthetic data across all metrics, indicating that tactile inputs provide useful local geometric cues even when the model is trained only on synthetic data and evaluated on real scenes. However, finetuning OmniDC further with real RGB and depth outperforms both synthetic-only models in most metrics, suggesting that real-domain fine-tuning is important for adapting to real sensor noise, object appearance, and boundary artifacts. 
Tactile-DC achieves the best overall performance across both general and thin-object focused evaluation. Under the default protocol for object evaluation, Tactile-DC improves over OmniDC finetuned on the full dataset, both real and synthetic, from 0.717/0.807/0.848 to 0.719/0.811/0.852 in A1--A3, while reducing AbsRel from 0.218 to 0.205, MAE from 0.082 to 0.078, and RMSE from 0.213 to 0.205. The gains are also consistent under linear-fit and median-scale protocols. In particular, Tactile-DC obtains the strongest median-scale A1 and RMSE on object evaluation, indicating best reconstruction quality on thin-object pixels. Overall, the comparison shows that training on synthetic data could already improve depth completion performance considerably despite being evaluated on real data. The improvement from further finetuning  models that were pretrained on synthetic data, on real data, shows that real-domain adaptation also boosts performance substantially, while Tactile-DC achieves the strongest results by combining this adaptation with local tactile constraints.

\begin{table*}[t!]
\centering
\caption{Evaluation metrics across NVS methods and protocols on the NVS test split of the real dataset.}
\resizebox{0.99\textwidth}{!}{
\begin{tabular}{l l r r r r r r r r} 
\toprule
\textbf{Model} & \textbf{Protocol} & \textbf{A1} & \textbf{A2} & \textbf{A3} & \textbf{AbsRel} & \textbf{logMAE} & \textbf{MAE} & \textbf{logRMSE} & \textbf{RMSE} \\
\midrule

\rowcolor{gray!15}
\multicolumn{10}{c}{\textbf{General Evaluation}} \\
\midrule

\multirow{3}{*}{2D Gaussian Splatting} 
  & Default & 0.013 & 0.026 & 0.040 & 2.054 & 0.497 & 1.239 & 1.560 & 1.784 \\
  & Linear Fit & 0.069 & 0.137 & 0.204 & 0.613 & 0.206 & 0.557 & 0.475 & 0.591 \\
  & Median Scale & 0.119 & 0.206 & 0.275 & 0.689 & 0.254 & 0.761 & 0.662 & 0.942 \\
\midrule

\multirow{3}{*}{3D Gaussian Splatting} 
  & Default & 0.013 & 0.025 & 0.038 & 2.019 & 0.491 & 1.222 & 1.673 & 2.047 \\
  & Linear Fit & 0.055 & 0.110 & 0.163 & 0.682 & 0.223 & 0.587 & 0.499 & 0.618 \\
  & Median Scale & 0.098 & 0.186 & 0.262 & 0.694 & 0.253 & 0.759 & 0.700 & 1.084 \\
\midrule

\multirow{3}{*}{Nerfacto} 
  & Default & 0.005 & 0.011 & 0.022 & 6.660 & 0.641 & 1.668 & 3.660 & 4.424 \\
  & Linear Fit & 0.052 & 0.104 & 0.157 & 0.778 & 0.238 & 0.645 & 0.552 & 0.676 \\
  & Median Scale & 0.086 & 0.164 & 0.223 & 2.762 & 0.471 & 1.311 & 1.540 & 2.356 \\
\midrule

\multirow{3}{*}{Depth-Nerfacto} 
  & Default & 0.006 & 0.014 & 0.027 & 3.267 & 0.480 & 1.225 & 2.544 & 3.418 \\
  & Linear Fit & 0.042 & 0.085 & 0.127 & 0.822 & 0.257 & 0.670 & 0.581 & 0.692 \\
  & Median Scale & 0.171 & 0.321 & 0.426 & 1.515 & 0.284 & 0.911 & 1.370 & 2.469 \\
\midrule

\rowcolor{gray!15}
\multicolumn{10}{c}{\textbf{Object Evaluation}} \\
\midrule

\multirow{3}{*}{2D Gaussian Splatting} 
  & Default & 0.005 & 0.010 & 0.014 & 2.740 & 0.557 & 1.301 & 1.320 & 1.366 \\
  & Linear Fit & 0.031 & 0.066 & 0.105 & 0.983 & 0.267 & 0.633 & 0.430 & 0.455 \\
  & Median Scale & 0.126 & 0.213 & 0.288 & 1.161 & 0.236 & 0.579 & 0.517 & 0.580 \\
\midrule

\multirow{3}{*}{3D Gaussian Splatting} 
  & Default & 0.007 & 0.013 & 0.020 & 2.697 & 0.550 & 1.292 & 1.303 & 1.363 \\
  & Linear Fit & 0.027 & 0.054 & 0.083 & 1.055 & 0.284 & 0.665 & 0.451 & 0.466 \\
  & Median Scale & 0.089 & 0.175 & 0.248 & 1.173 & 0.248 & 0.615 & 0.513 & 0.595 \\
\midrule

\multirow{3}{*}{Nerfacto} 
  & Default & 0.009 & 0.022 & 0.050 & 2.193 & 0.376 & 0.972 & 0.941 & 1.583 \\
  & Linear Fit & 0.010 & 0.021 & 0.032 & 1.547 & 0.375 & 0.875 & 0.683 & 0.695 \\
  & Median Scale & 0.051 & 0.128 & 0.205 & 0.860 & 0.392 & 0.962 & 0.396 & 0.725 \\
\midrule

\multirow{3}{*}{Depth-Nerfacto} 
  & Default & 0.009 & 0.024 & 0.053 & 1.455 & 0.338 & 0.823 & 0.605 & 0.879 \\
  & Linear Fit & 0.010 & 0.022 & 0.036 & 1.164 & 0.309 & 0.719 & 0.513 & 0.519 \\
  & Median Scale & 0.192 & 0.397 & 0.529 & 0.443 & 0.186 & 0.493 & 0.196 & 0.428 \\

\bottomrule
\end{tabular}}
\label{tab:nvs_depth_est_real}
\end{table*}

\begin{figure}[h!]
    \centering
    \includegraphics[width=0.9\linewidth]{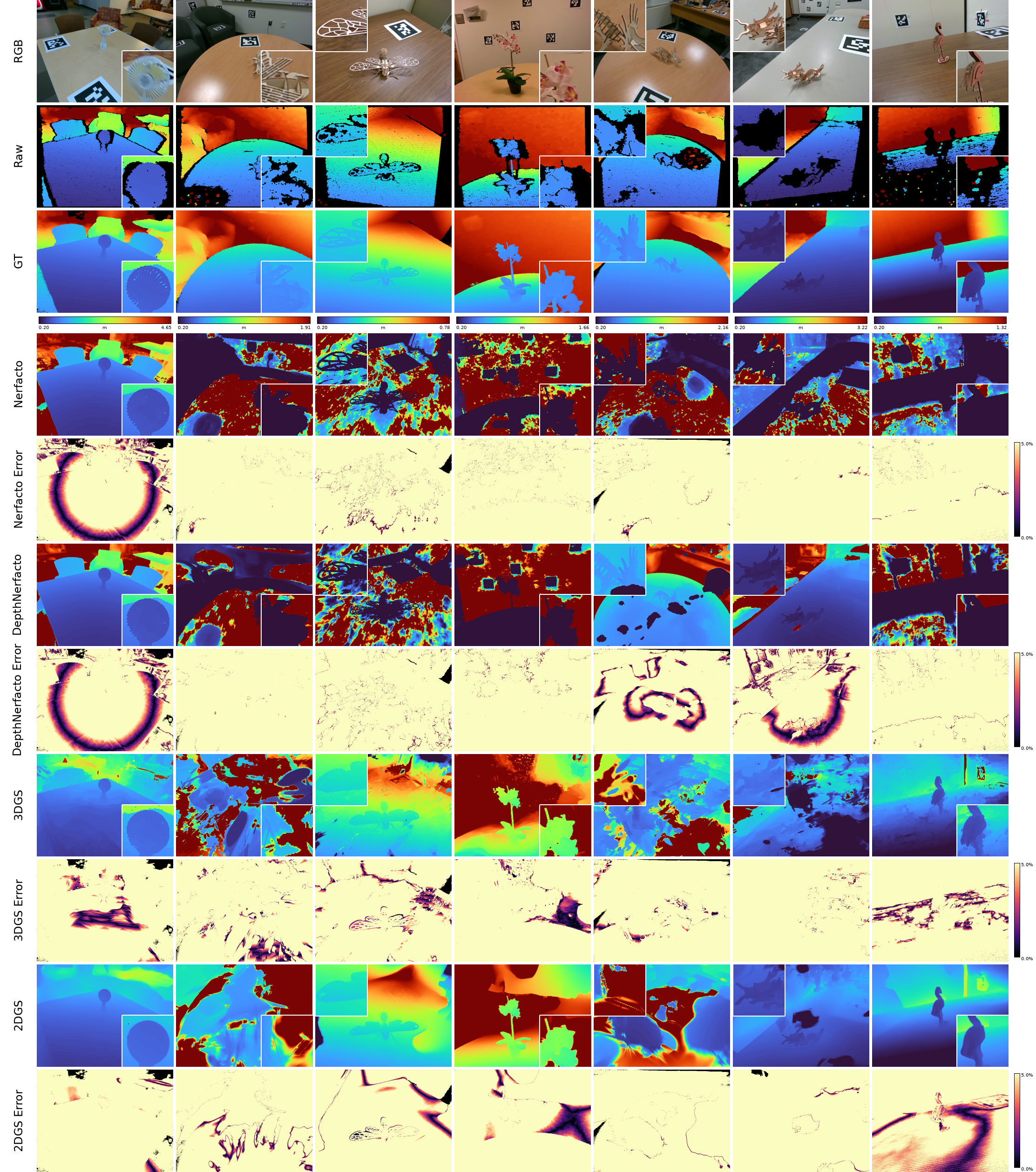}
    \caption{Comparison of the RGB image, raw sensor depth, ground truth depth, and depth estimates from Novel-view-synthesis baselines. All baseline depth estimates are median-scaled with respect to the ground truth. 
}
    \label{fig:nvs_depth_error}
\end{figure}

\textbf{Depth Estimation Evaluation on Novel View Synthesis (NVS) Methods.}
Table~\ref{tab:nvs_depth_est_real} evaluates the depth maps recovered from NVS methods on the real NVS test split, and Fig.~\ref{fig:nvs_depth_error} provides qualitative comparisons. Overall, all NVS methods perform poorly under the default protocol, with A1--A3 close to zero, indicating that good view synthesis does not necessarily produce metrically accurate depth. Linear fitting and median scaling improve performance, showing that part of the error comes from global scale mismatch, but the remaining low object accuracy indicates difficulty on thin-object regions. 

Among Gaussian-splatting methods, 2D Gaussian Splatting slightly outperforms 3D Gaussian Splatting after scale alignment, especially under object evaluation, but both still exhibit strong errors and fragmented thin structures in Fig.~\ref{fig:nvs_depth_error}. For NeRF-based methods, Depth-Nerfacto improves over Nerfacto in most settings and achieves the best median-scale results for object evaluation, suggesting that depth supervision helps performance on recovering scene structure when scale is aligned. However, its results still exhibit significant scale misalignment issues even with median scaling. Overall, NVS methods remain unreliable for accurate thin-object depth reconstruction.

\section{Visualization of Depth Distribution}

\begin{figure*}[t]
    \centering
    \includegraphics[width=0.99\linewidth]{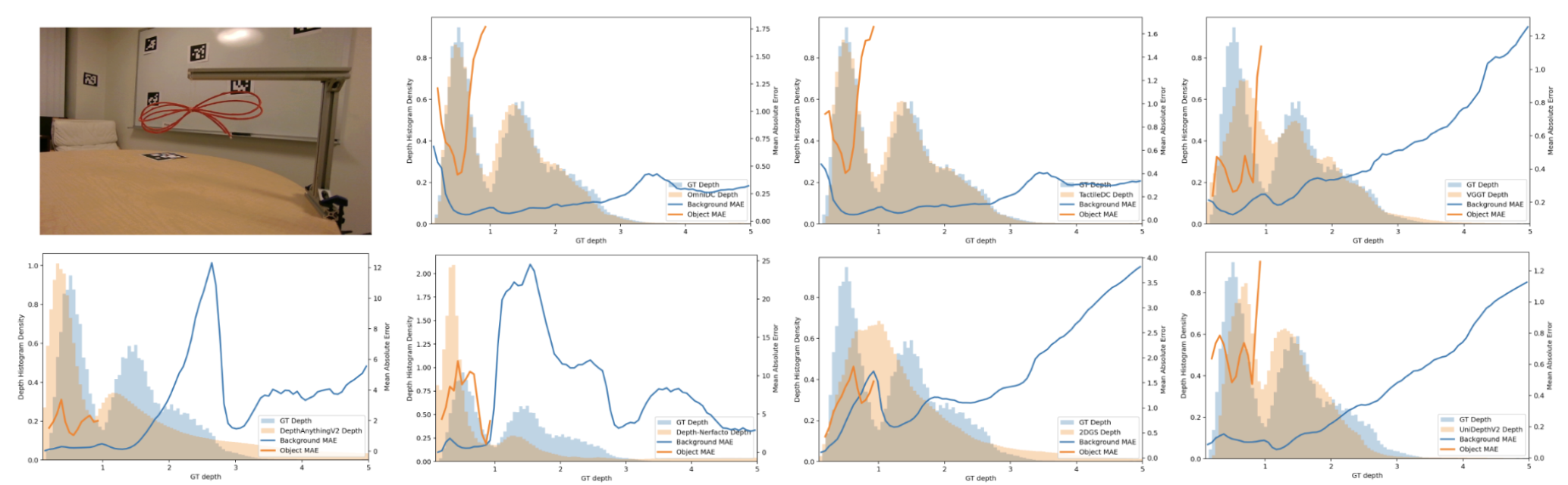}
    \caption{Depth distribution plotted representative methods. Blue line represents MAE of depth predictions for all ranges in depth; orange line presents MAE of depth predictions on thin objects; orange histogram represents the distribution of depth predictions by Tactile-DC and baselines; blue histogram represents the GT depth distribution.}
    \label{fig:characteristic_plots}
\end{figure*}

We show characteristic plots for Tactile-DC and representative baselines on a representative scene in Fig. \ref{fig:characteristic_plots}, highlighting differences in the predicted depth distributions. The lines corresponding to the object exhibit higher error than the average error over the entire scene. This demonstrates that thin object regions are more challenging to reconstruct for 3D reconstruction models. Furthermore, the plots show that Tactile-DC produces a depth distribution closer to the ground-truth distribution than other baselines. 

\section{Depth Estimation Metrics}
\label{sec:mde-metrics}
To evaluate depth estimation models on the dataset, we employ metrics commonly used in prior works ~\cite{eigen2014depth, fu2018deep}. Metrics include threshold accuracy $\delta < 1.05$, absolute relative error (AbsRel), logarithmic mean absolute error (logMAE), mean absolute error (MAE), logarithmic root mean squared error (logRMSE), and root mean squared error (RMSE). Higher values are better in threshold accuracy, while lower values are better in other metrics.

Let $d(x)$ denote the ground-truth depth at pixel $x$, $\hat{d}(x)$ the predicted depth, and $\Omega$ the set of valid pixels with available ground-truth depth.

\textbf{Threshold Accuracy ($\delta < 1.05$):}
\begin{equation}
    \delta < 1.05 =
    \frac{1}{|\Omega|}
    \sum_{x \in \Omega}
    \mathds{1}\left(
    \max\left(
    \frac{\hat d(x)}{d(x)}, 
    \frac{d(x)}{\hat d(x)}
    \right) < 1.05
    \right).
\end{equation}

\textbf{Absolute Relative Error (AbsRel):}
\begin{equation}
\text{AbsRel} =
\frac{1}{|\Omega|}
\sum_{x \in \Omega}
\frac{|\hat d(x) - d(x)|}{d(x)}.
\end{equation}

\textbf{Mean Absolute Error (MAE):}
\begin{equation}
\text{MAE} =
\frac{1}{|\Omega|}
\sum_{x \in \Omega}
\left| \hat d(x) - d(x) \right|.
\end{equation}

\textbf{Logarithmic Mean Absolute Error (logMAE):}
\begin{equation}
\text{logMAE} =
\frac{1}{|\Omega|}
\sum_{x \in \Omega}
\left|
\log \hat d(x) - \log d(x)
\right|.
\end{equation}

\textbf{Root Mean Squared Error (RMSE):}
\begin{equation}
\text{RMSE} =
\left(
\frac{1}{|\Omega|}
\sum_{x \in \Omega}
\left(\hat d(x) - d(x)\right)^2
\right)^{1/2}.
\end{equation}

\textbf{Logarithmic Root Mean Squared Error (logRMSE):}
\begin{equation}
\text{logRMSE} =
\left(
\frac{1}{|\Omega|}
\sum_{x \in \Omega}
\left(
\log \hat d(x) - \log d(x)
\right)^2
\right)^{1/2}.
\end{equation}

\section{Perception-based Metrics}
\label{sec:perception-metrics}
To evaluate the quality of the synthesized novel view by NVS methods on our dataset, we utilize the image quality metrics

\noindent\textbf{Peak Signal-to-Noise Ratio (PSNR)} measures the difference between the reference image and the synthesized image across all three color channels. While the Mean Squared Error (MSE) is calculated by the mean of the squared differences over RGB channels, the peak signal-to-noise ratio is defined by:

\begin{equation}
PSNR = 10 \cdot \log_{10}\left(\frac{{MAX_I}^2}{MSE}\right),
\end{equation}
where the $MAX_I$ is the maximum possible pixel value.

\noindent\textbf{Structural Similarity Index Measure (SSIM)} is used to measure the changes in structures.

Given the reference image $I_{\text{ref}}$ and the synthesized image $I_{\text{syn}}$, and their statistics (i.e. $\mu_x$, the pixel mean of an image $x$, $\sigma_x^2$, the sample variance of image $x$, and $\sigma_{x,y}$, covariance of image $x$ and $y$), SSIM is defined by 
\begin{equation}
    SSIM(I_{\text{ref}}, I_{\text{syn}})= \frac{(2\mu_{\text{ref}} \cdot \mu_{\text{syn}}+c_1)(2\sigma_{\text{ref,syn}} + c_2 )}{(\mu_{\text{ref}}^2 + \mu_{\text{syn}}^2+c_1)(\sigma_{\text{ref}}^2+\sigma_\text{syn}^2 + c_2)},
\end{equation}
where $c_1$ and $c_2$ are the predefined constant values.

\noindent\textbf{Learned Perceptual Image Patch Similarity (LPIPS)}~\cite{zhang2018unreasonable} is utilized to measure the perceptual difference by deep neural network features, which computes the patch-wise similarity between the activations of two images.

\section{Real-World Dataset Details}
\subsection{Object Assets}

\begin{figure}[t]
    \centering
    \includegraphics[width=0.99\linewidth]{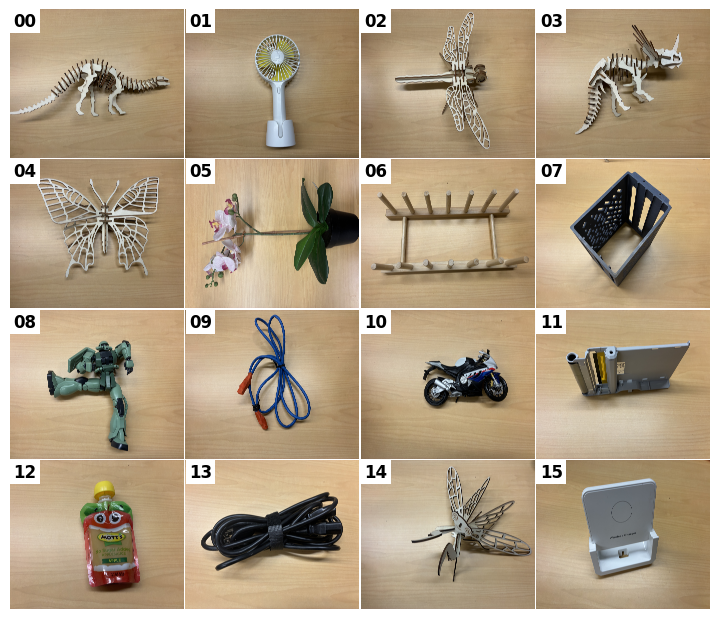}
    \caption{Part of the real-world objects used in the dataset. The numbers in each figure indicate the ID.}
    \label{fig:real_objects_examples}
\end{figure}

See Fig.~\ref{fig:real_objects_examples} for representative examples of our real-world objects.

\subsection{Scene Selection}
We captured the VisTa3D dataset in eight different background environments with diverse lighting conditions. The first scene is a well-lit conference room with a semi-reflective table, with minimal background clutter, providing clean scene structure while introducing reflection artifacts from the tabletop. The second scene is a dim hallway, where tighter spatial constraints and reduced illumination create more challenging visual conditions. The third scene is a compact, cluttered office with a near-distance background depth. The fourth scene is a well-illuminated kitchen scene with countertops and cabinets around the wall. The fifth scene is an outdoor bench under a roof, illuminated by indirect sunlight; this setting introduces natural lighting variation, and the close camera range captures fine details of the thin objects. The sixth scene is a dim cubicle with limited lighting, minimal background clutter and complex camera trajectory. The seventh scene is a very dim lobby with abundant background objects and a wide depth range, introducing challenges from low-contrast, background clutter, and depth variation. The eighth scene is an outdoor staircase under direct sunlight, with strong illumination changes, shadows, and the unique structure of the staircase. Together, these settings provide a diverse range of lighting conditions, scene layouts, and visual challenges for evaluating thin-object reconstruction.

\subsection{Post-Processing}
\label{subsec:mono_filtering}
The background of the back-projected point cloud from depth maps contains artifacts and erroneous estimates, particularly in distant regions.
We therefore apply the following background-specific post-processing methods to mitigate the background depth noise while preserving near-field structure.

\textbf{Local Support Surface Refinement.} 

Because the target object is in contact with the surrounding scene structure, the local surface provides a useful prior for separating $P_{\text{obj}}$ from $P_{\text{raw}}$. Here, we denote the remaining component of $P_{\text{raw}}$ as $P_{\text{background}}$. However, view-dependent appearance changes (e.g. reflectance) and depth estimation failures near contact region can introduce floating artifacts in the $P_{\text{background}}$. 

To address this issue in a scene-agnostic manner, we perform local surface refinement around the estimated object region. We first identify the neighborhood of $P_{\text{obj}}$ and use the nearby background structure as a local support prior. Points within this region are filtered according to their consistency with the surrounding structure: points that deviate from the local surfaces are removed, while structurally consistent background points are retained. This cleanup process suppresses artifacts near the object, smooths the adjacent surfaces, and avoids damaging the scene. The refined point cloud therefore preserves background structure and provides a cleaner representation for subsequent re-projection. 

\textbf{Robust Depth Outlier Removal.} 
In addition to applying the general statistical outlier removal (SOR)~\cite{rusu20113d}, we also adopt a light voxel-based connected-component (small-cluster) filtering to remove small floating clusters that SOR struggles to handle. The voxelized background point cloud $P_{\text{background}}$ forms a sparse grid, builds connected components in this voxel space using 6/26-neighborhood adjacency, and then discards components whose size falls below a threshold. The remaining points are kept and mapped back to the original resolution, which reduces isolated fragments while preserving the main object and large planar structures.

We also conduct an edge-focused outlier removal, since raw depth is often unreliable near object boundaries. The method detects small, local depth patches that are inconsistent with their immediate surroundings by comparing the average depth of each patch to that of a narrow ring around it. Patches with large depth discrepancies are removed, effectively filtering out trailing edge clusters while preserving locally consistent surfaces.

\subsection{Implementation Details}
For all methods except Tactile-DC, we use the authors' open-source implementations and pretrained checkpoints when available, and modify only data paths, input formatting, and image resolution to match our protocol. Each method is provided with the subset of available inputs required by its original formulation. For all NVS methods, the ORB\_SLAM3 poses are given. Sparse depth maps are generated by detecting FAST keypoints in each RGB frame, randomly sampling 1,500 valid keypoint locations, and retaining the corresponding metric depth values from the raw sensor depth while setting all other pixels to zero. For MapAnything, we use the ``Images + Calibration + Depth'' mode and provide the sparse depth maps as input depth. OmniDC and Tactile-DC also use sparse depth as part of their inputs, while Depth-Nerfacto uses the raw sensor depth maps. Due to memory constraints, VGGT uses sliding windows of 30 frames over each sequence as input views. Since Tactile-DC reuses components of standard depth completion models, we initialize the model parameters with pretrained weights. As the tactile modulation component of our method is novel, its parameters are trained from scratch. We trained the model on the VisTa3D training set for 20 epochs with standard photometric and geometric augmentations of the RGB images, sparse depth maps, and tactile response maps following AugUndo \cite{wu2024augundo}.

\section{Synthetic Dataset Details}

\subsection{Tabletop}

\begin{figure}[t]
    \centering
    \includegraphics[width=0.9\linewidth]{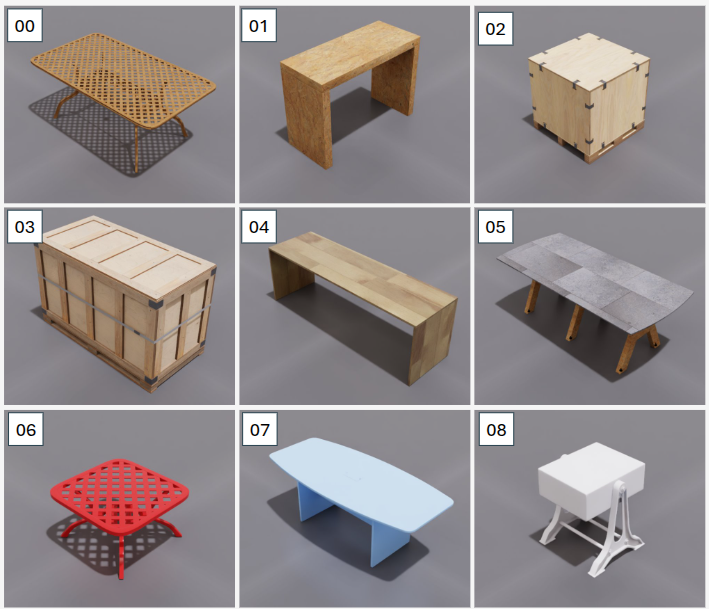}
    \caption{The tabletops used in the synthetic dataset. The numbers in each figure indicate the ID.}
    \label{fig:tabletop_examples}
\end{figure}

We selected nine tabletop assets from the SimReady Explorer within NVIDIA Isaac Sim to introduce domain randomization and increase structural and visual diversity. As shown in Fig. \ref{fig:tabletop_examples}, the assets include a range of surface types such as wood tables, flooring-like surfaces, and perforated surfaces. These variations help to broaden the distribution of background appearances. A list of the nine tabletop assets used in our scenes is provided in Table \ref{tab:tabletop}.

\begin{table}[h!]
\centering
\caption{Tabletop Used in the Synthetic Dataset}
\label{tab:tabletop}
\begin{tabular}{c|l}
\hline
\textbf{ID} & \textbf{Tabletop Name} \\ \hline
00 & Dellwood\_diningtable \\
01 & Danny \\
02 & Heavyduty\_wood\_crate\_assembly \\
03 & Plywood\_crate\_assembly \\
04 & Oak\_table\_large \\
05 & Cline \\
06 & Dellwood\_endtable \\
07 & Boat \\
08 & Appleseed\_endtable \\
\hline
\end{tabular}
\end{table}

\begin{figure*}[t]
    \centering
    \includegraphics[width=0.9\linewidth]{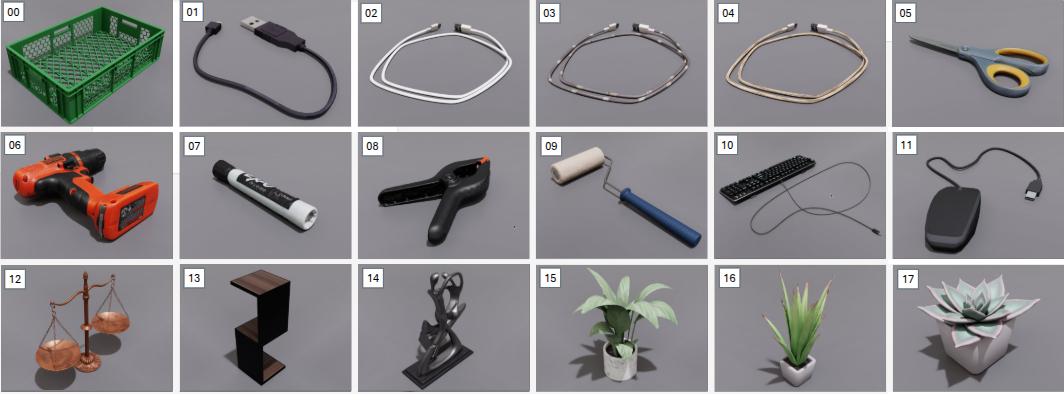}
    \caption{The objects used in the synthetic dataset. The numbers in each figure represent the ID. Object ID 02 was retextured to create versions 03 and 04.}
    \label{fig:object_examples}
\end{figure*}

\subsection{Object Assets and Modifications}

We selected the 18 object assets from a combination of sources, including the built-in NVIDIA Isaac Sim asset library, Sketchfab 3D models, and several customized wire-based objects that we created ourselves. Similarly to the tabletop configurations, some objects are also recolored to introduce additional appearance diversity. As shown in Fig. \ref{fig:object_examples}, it can be observed that objects contain at least one thin part or wire-based object. A complete list of 18 object assets used in our scenes is provided in Table \ref{tab:objects_full}.

\begin{table*}[t]
\centering
\caption{Complete list of 18 object assets used in the synthetic dataset, including brief descriptions and object type. Object type refers to state information such as whether it is rigid or non-rigid when simulated. }
\label{tab:objects_full}
\resizebox{0.99\textwidth}{!}{\begin{tabular}{lllll}
\toprule
\textbf{ID} & \textbf{Object Model Name} & \textbf{Description} & \textbf{Object type} & \textbf{Source}\\
\midrule
\multicolumn{5}{l}{\textbf{Rigid Objects}} \\
00 & Paint\_roller & Cylindrical paint roller tool& Rigid & Isaac sim asset library \\
01 & Container& Plastic rectangular storage container & Rigid & Isaac sim asset library \\
02 & USB\_A\_USB\_B & USB cable with connectors on both ends & Rigid  & Sketchfab\\
03 & USB\_C & Type-C connector cable & Rigid & Sketchfab \\
04 & USB\_C & Type-C connector cable & Rigid & Sketchfab \\
05 & USB\_C & Type-C connector cable & Rigid & Sketchfab \\
06 & Large\_marker & Cylindrical marker pen & Rigid  & Isaac sim asset library\\
07 & Large\_clamp & Plastic clamp & Rigid  & Isaac sim asset library\\
08 & Wired\_keyboard & Full-size wired keyboard & Rigid  & Sketchfab\\
09 & Wired\_mouse & Wired computer mouse & Rigid  & Sketchfab\\
10 & Corner\_shelf & Corner shelf structure & Rigid  & Isaac sim asset library\\
11 & Dancer & Decorative thin sculpture & Rigid  & Isaac sim asset library\\
12 & Plant & Potted small indoor plant & Rigid  & Isaac sim asset library\\
13 & Plant & Potted medium plant & Rigid & Isaac sim asset library \\
14 & Plant & Potted tall plant & Rigid  & Isaac sim asset library\\
15 & Copper\_scales & Small metallic weight scale & Rigid & Isaac sim asset library \\
16 & Power\_drill & Hand-held power tool & Rigid & Isaac sim asset library \\
17 & Scissors & Metal scissors with thin blades & Rigid & Isaac sim asset library \\
\end{tabular}
}
\end{table*}

\subsection{Scene Generation}

\begin{figure}[t]
    \centering
    
    \begin{subfigure}[b]{0.48\linewidth}
        \centering
        \includegraphics[width=\linewidth]{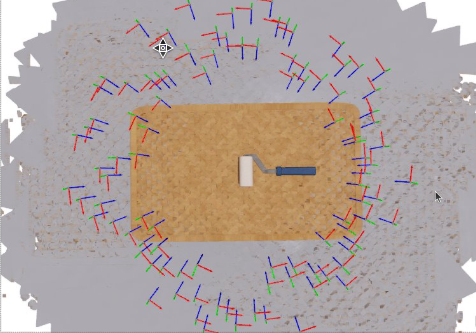}
        \caption{}
        \label{fig:a}
    \end{subfigure}
    \hfill
    \begin{subfigure}[b]{0.48\linewidth}
        \centering
        \includegraphics[width=\linewidth]{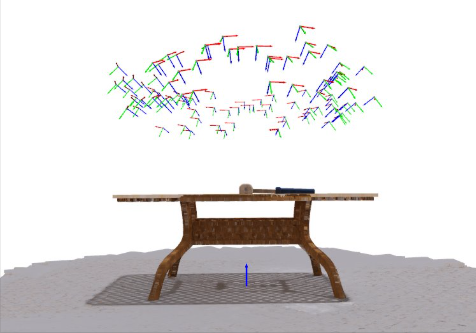}
        \caption{}
        \label{fig:b}
    \end{subfigure}

    \caption{The Paint roller on the Dellwood diningtable. (a) Top view of the scene. (b) Sagittal view of the scene. In (a), all frames correspond to the camera frame, whereas in (b), the lower frame represents the world frame and the upper frame represents the camera frame.}
    \label{fig:synthetic_scene_views}
\end{figure}

As mentioned in the main paper, we generated the data with a camera rotating 360 degrees around the object on tabletop configuration. To avoid fixed camera elevation and radial distance that would result in monotonous data or simple turntable-style data, we added uniform noise. This additive noise consisted of random values to the radius by up to 0.2 m and the camera height by up to 0.3 m. As shown in Fig. \ref{fig:synthetic_scene_views}, the camera position was captured in a way that was not a monotonous orbit. This configuration provides a variety of viewpoints of the data. Data generation was conducted using NVIDIA Isaac Sim 5.0 on a system equipped with an NVIDIA RTX 4090 GPU and an Intel i9-13900 CPU.

\section{Extended Discussion on Future Works}

VisTa3D is the first multimodal dataset and benchmark for thin object 3D reconstruction. While we primarily focus on thin objects, there exist many challenges that 3D reconstruction models face including highly specular, reflective, and transparent objects. Our data collection platform is limited in that the sensors cannot reliably capture such objects, and hence they were omitted from the dataset. However, our methodology has the potential to be extended to these challenging objects, by registering known object shapes to them. We also focused on single object data collection. We note that a subset of our objects is deformable. However, they have been tied and made static. This is a potential opportunity for dynamic 3D reconstruction \cite{wang2026ode,zhang2025monst3r} or other motion estimation problems \cite{lao2024diffeomorphic,lao2018extending,xiao2026triangular,zhang2024adaptive,zhang2024heteroscedastic}; we leave this for future work. Additionally, a natural extension of our dataset would be to consider multiple objects, this will introduce more clutter and occlusion, which will further pose challenges for 3D reconstruction models. We also focused only on a standard perspective camera in our capture platform. It is possible to extend our camera setup to consider wide-angle or fisheye lens \cite{duan2026fisheye3r,gangopadhyay2025extending,gangopadhyay2026from,lichy2024fova,guo2025depth}, which are commonly used in spatial applications. Furthermore, inclusion of additional modalities, such as text, may allow for the benchmarking of tactile understanding \cite{yang2024binding,tu2026unitac,zhu2025forces}.

We tested supervised methods in monocular depth estimation \cite{lao2024depth,lao2024sub,piccinelli2025unidepthv2,yang2024depthanythingv2}, monocular depth completion \cite{zuo2025omni,ezhov2024all,rim2026radar,singh2023depth}, multi-view stereo \cite{wang2025vggt}, and neural rendering methods \cite{duan2026evidential,kerbl20233d,mildenhall2021nerf}. For future work we will also consider unsupervised methods, such as those in monocular depth estimation \cite{fei2019geo,garg2016unsupervised,godard2017unsupervised,godard2019digging,wong2019bilateral} and depth completion \cite{liu2022monitored,park2026orcas,rim2025protodepth,wong2020unsupervised,wong2021adaptive,wong2021learning,wong2021unsupervised} as well as test-time adaptive methods \cite{chung2025eta,park2024test,zhang2025progressive} to gauge whether error modes can be mitigated through model updates.

\clearpage

\end{document}